**RESEARCH ARTICLE**

# A Semi-Automated Solution Approach Recommender for a Given Use Case: a Case Study for AI/ML in Oncology via Scopus and OpenAI

Deniz Kenan Kılıç[1] · Alex Elkjær Vasegaard[1] · Aurélien Desoeuvres[1] · Peter Nielsen[1]

## Abstract
Nowadays, literature review is a necessary task when trying to solve a given problem. However, an exhaustive literature review is very time-consuming in today's vast literature landscape. It can take weeks, even if looking only for abstracts or surveys. Moreover, choosing a method among others, and targeting searches within relevant problem and solution domains, are not easy tasks. These are especially true for young researchers or engineers starting to work in their field. Even if surveys that provide methods used to solve a specific problem already exist, an automatic way to do it for any use case is missing, especially for those who don't know the existing literature. Our proposed tool, SARBOLD-LLM, allows discovering and choosing among methods related to a given problem, providing additional information about their uses in the literature to derive decision-making insights, in only a few hours. The SARBOLD-LLM comprises three modules: (1: Scopus search) paper selection using a keyword selection scheme to query Scopus API; (2: Scoring and method extraction) relevancy and popularity scores calculation and solution method extraction in papers utilizing OpenAI API (GPT 3.5); (3: Analyzes) sensitivity analysis and post-analyzes which reveals trends, relevant papers and methods. Comparing the SARBOLD-LLM to manual ground truth using precision, recall, and F1-score metrics, the performance results of AI in the oncology case study are 0.68, 0.9, and 0.77, respectively. SARBOLD-LLM demonstrates successful outcomes across various domains, showcasing its robustness and effectiveness. The SARBOLD-LLM addresses engineers more than researchers, as it proposes methods and trends without adding pros and cons. It is a useful tool to select which methods to investigate first and comes as a complement to surveys. This can limit the global search and accumulation of knowledge for the end user. However, it can be used as a director or recommender for future implementation to solve a problem.

## Highlights

- Automated support for literature choice and solution selection for any use case.
- A generalized keyword selection scheme for literature database queries.
- Trends in literature: detecting AI methods for a case study using Scopus and OpenAI.
- A better understanding of the tool by sensitivity analyzes for Scopus and OpenAI.
- Robust tool for different domains with promising OpenAI performance results.

✉ Deniz Kenan Kılıç
denizkk@mp.aau.dk

Alex Elkjær Vasegaard
aev@mp.aau.dk

Aurélien Desoeuvres
aureliend@mp.aau.dk

Peter Nielsen
peter@mp.aau.dk

1 Department of Materials and Production, Aalborg University, Fibigerstræde 16, 9220 Aalborg, Denmark

## 1 Introduction

Over the past decade, artificial intelligence (AI) and machine learning (ML) have gained significant attention in the fields of information technology and computer science, accompanying significant advancements and benefits across diverse industries and sectors [1, 2]. There are numerous AI/ML taxonomies presented in the literature [3, 4] that can be used to select a collection of AI strategies to address a specific

Springer

challenge.[1] Figure 1 illustrates an example taxonomy of the extensive AI/ML domain, encompassing multiple problem types and branches. However, to search for AI methods specific to a given use case, it is not only necessary to select a fitting branch in the taxonomy, but one also has to refine the search by comparing it to the standing knowledge base of the literature on the use case.

The increasing amount of literature presents a challenge for decision-makers seeking to employ AI/ML methodology in their specific problem domains. Manual review is time-consuming [5], often resulting in incomplete information without targeted searches. The current way to reduce the time spent in choosing a method consists of reading reviews or surveys that consider and explain the pros and cons of several methods belonging to a given field. This has however some limitations, a review considers in general a family of comparable and well-known methods but cannot provide an exhaustive list. Moreover, they do not consider the uses or interests of these methods over time. A tool that rapidly generates trend findings and examines solution methods for any use case would be extremely beneficial in various situations.

This research proposes a semi-automated tool developed to generate results on solution approaches for any use case. The tool is named SARBOLD-LLM where SARBOLD stands for Solution Approach Recommender Based On Literature Database and LLM stands for Large Language Model, respectively. Considering all the constraints and preferences needed in the use case, the objective is to help the end user provide a list of methods able to solve a given problem (the use case), sort them in such a way that it is easy for the end user to choose a method to investigate first and supply existing literature regarding this chosen method.

The study presents results on multiple problem domains on AI with a focus on the case study for AI/ML in oncology. The SARBOLD-LLM has three modules called "Module 1: Scopus search", "Module 2: Scoring and method extraction", and "Module 3: Analyzes", respectively. It broadly contains the following steps:

- Determining keywords systematically from the use case by a two-domain, three-level setup. (Module 1)
- Automated literature extraction using selected keywords via Scopus Search application programming interface (API) [6]. (Module 1)
- Extracting AI methods automatically from Scopus search results by using OpenAI API (text-davinci-003, GPT 3.5). (Module 2)

---
[1] https://scikit-learn.org/stable/tutorial/machine_learning_map/index.html

- Sensitivity analyzes for both Scopus and OpenAI. (Module 3)
- Post-analyzes based on results. (Module 3)

Some of the existing studies, which work on solution approach suggestions, have worked on the reduction of time and effort spent and automation. In addition to these, SARBOLD-LLM uses a keyword selection strategy to make sure the search is inside the relevant problem and solution domains. Additionally, it uses trend, popularity, and relevancy analyzes to draw decision-making conclusions about the solution techniques used for any given use case. Furthermore, the sensitivity analyzes performed for Scopus and OpenAI and the performance results obtained for OpenAI are among the positive values of this research.

The SARBOLD-LLM can be used iteratively for the decision makers to augment their understanding of the problem and similarly align the keywords better with the desired use case and specificity level, consequently obtaining better results.

The remainder of this article is structured as follows: Section 2 reviews the use of AI methods and the literature on model selection approaches. Section 3 presents the SARBOLD-LLM, and Section 4 showcases the performance, sensitivity, and post-analysis of the method. In Section 5, a discussion is given. Finally, a conclusion and suggestions for future works are provided in Section 6.

## 2 Literature Review

In the literature, there are reviews and surveys on which AI approaches or applications are used for different problem domains such as building and construction 4.0 [7], architecture, engineering and construction (AEC) [8], agriculture [9], watermarking [10], healthcare [11, 12], oil and gas sector [13], supply chain management [14], pathology [15], banking [16], finance [17], food adulteration detection [18], engineering and manufacturing [19], renewable energy-driven desalination system [20], path planning in the unmanned aerial vehicle (UAV) swarms [21], military [22], cybersecurity management [23], engineering design [24], vehicular ad-hoc networks [25], dentistry [26], green building [27], e-commerce [28], drug discovery [29], marketing [30], electricity supply chain automation [31], monitoring fetus via ultrasound images [32], internet of things (IoT) security [33]. In Table 1, AI approaches utilized in different problem domains are illustrated.

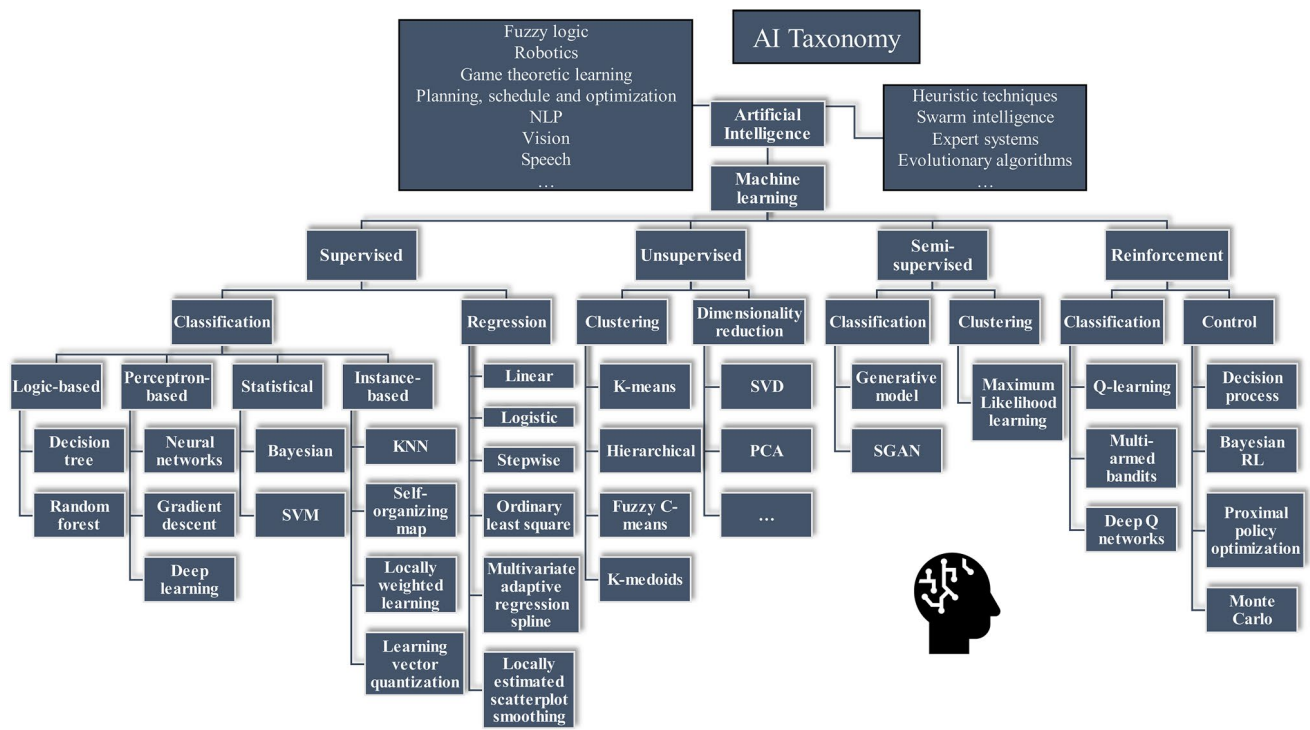

**Fig. 1** An example of AI/ML taxonomy

As can be seen from the aforementioned references, some of the problem domains in the example review and surveys are low-level, while some are high-level. The abstraction level is difficult to integrate for the solution domain while considering the reviews and surveys. Even if the same problem domain is considered, it will be an issue to depend on reviews or surveys in the literature as there may be an unlimited number of use case scenarios and levels of specificity. In addition, AI approaches specified in reviews or surveys can sometimes be very general. In this case, it may be necessary to make article reviews manually, but it causes labor and time loss [52]. Based on this idea, one can search for an automated way to minimize the time spent on manual review to get an AI method applied to a given use case.

The last decade saw significant steps toward a fully automatic model selection scheme with tools that select models for specialized use cases, generally referred to as model determination, parameter estimation, or hyper-parameter selection tools. For forecasting time series in R, the popular *forecast* package by R. Hyndman et al. was presented, showcasing great initial results [53]. For regression models, the investigated selection procedures are generally based on the evaluation of smaller pre-defined sets of alternative methods, e.g., by information criteria (AIC, BIC), shrinkage methods (Lasso), stepwise regression, and or cross-validation schemes [54]. For ML-based model schemes, the methods proposed by B. Komer et al. [55] introduce the *hyperopt* package for hyper-parameter selection accompanying the Scikitlearn ML library, J. Snoek et al. [56] presents a Bayesian optimization scheme to identify the hyper-parameter configuration efficiently, and J. Bergstra et al. [57] identifies hyper-parameter configurations for training neural networks and DBNs by using a random search algorithm and two greedy sequential methods based on the expected improvement criterion. There also exist smaller frameworks, e.g., that of hyper-parameter tuning based on problem features with MATE [58], to model and fit autoregressive-to-anything processes in Java [59], or extensions to general-purpose optimization frameworks [60].

On the other hand, Dinter et al. [5] present a systematic literature review on the automation of systematic literature reviews with a concentration on all systematic literature review procedures as well as NLP and ML approaches. They stated that the main objective of automating a systematic literature review is to reduce time because human execution is costly, time-consuming, and prone to mistakes. Furthermore, the title and abstract are mostly used as features for several steps in the systematic review process proposed by Kitchenham et al. [61]. Even though our research does not stick to these procedures since our study was not a pure systematic literature review, the title and abstract are included for the OpenAI part. Additionally, they found the majority of systematic literature reviews to be automated using SVM and Bayesian networks, such as Naive Bayes classifiers, and there appears to be a distinct lack of evidence regarding the effectiveness of deep learning approaches in this regard.

**Table 1** AI approaches used in different problem domains

| Reference | Problem domain | AI solution approaches |
|---|---|---|
| [34] | Pneumonia image detection | convolution neural network (CNN), deep convolutional neural network (DCNN), k-nearest neighbor (KNN), RESNET, CheXNet, DECNET, artificial neural network (ANN), ResNet-101, ResNet-50, AlexNet, VGG-Net |
| [35] | Online spam review detection | support vector machine (SVM), bidirectional encoder representations from transformers (BERT), random forest, linear regression, eXtreme Gradient Boosting (XGBoost), generative adversarial network (GAN), long short-term memory (LSTM), bidirectional LSTM (BiLSTM), deep neural network (DNN), CNN, graph neural network (GNN), graph convolutional network (GCN) |
| [36] | Credit card fraud detection | ANN, federated learning, random forest, SVM, KNN, multiple-layer perception (MLP), XGBoost |
| [37] | Diabetes detection | neural network, KNN, SVM, logistic regression, random forest, reinforcement learning, decision tree |
| [38] | Crack detection for bridge infrastructure maintenance | scale-invariant feature transform (SIFT), oriented FAST and rotated BRIEF (ORB), speeded up robust features (SURF), CNN, GAN, fully convolution network (FCN), deeply supervised nets (DSN), SegNet |
| [39] | Place recognition for moving robot | principal component analysis (PCA), auto-encoder, DNN, SVM, latent Dirichlet allocation (LDA), back propagation neural network (BPNN), CNN, AMOSNet, HybridNet |
| [40] | Road network detection | U-Net, deep ResUNet, SVM, restricted Boltzmann machine (RBM), deep learning |
| [41] | Social relationship link inference | deep learning, GCN |
| [42] | Chronic kidney disease | K-means, XGBoost, neural network, random forest, SVM, random tree, bagging tree model (BTM), KNN, logistic regression, MLP, Naive Bayes, decision tree |
| [43] | Mental stress detection on Reddit posts | natural language processing (NLP), BERT, term frequency-inverse document frequency (TF-IDF), Word2Vec, deep learning, neural network, random forest, SVM, logistic regression, bag of words (BoW), kernel-based canonicalization network (KCNet), Naive Bayes, support vector classifier (SVC), LDA |
| [44] | Remaining useful life in mechanical systems | ANN, LSTM, recurrent neural network (RNN), CNN, GCN, random forest, support vector regression (SVR), BPNN, deep learning |
| [45] | Food image recognition | CNN, SVM, random forest, transfer learning, InceptionV3, VGG16, VGG19, Resnet, LSTM, RNN, Alexnet, Cafenet, DenseNet201 |
| [46] | Detecting fake news articles | CNN, BiLSTM, TF-IDF, Naive Bayes, DNN, transformer-based language models, BoW, LSTM, SVC, K-means, LSTM |
| [47] | Prediction of language and cognition rehabilitation outcomes of post-stroke patients | ElasticNet, random forest, KNN, XGBoost, logistic regression, ridge classifier, SVC, SVM, SVR, DNN, Bernoulli Naive Bayes, decision tree |
| [48] | Human activity recognition with location independence | LSTM, CNN, RNN, gated recurrent unit (GRU), AdaBoost, reinforcement learning |
| [49] | Future food production prediction | DenseNet, LSTM, ANN, deep belief network (DBN), SVM, MLP, CNN |
| [50] | Street object detection | GAN, CNN, region-based CNN (R-CNN), fast R-CNN, faster R-CNN, single shot detector (SSD), MobileNet SSD, SVM, you only look once (YOLO) |
| [51] | Cancer prediction | deep learning, CNN, RNN, transfer learning, genetic algorithm, SVM, decision tree, random forest, KNN, logistic regression, ANN, K-means, density-based spatial clustering of applications with noise (DBSCAN) |

**Table 2** Solution approach selection comparison table

| Methods / Features | Manual literature review via articles | Manual literature review via surveys and reviews | Automated literature review | SAR-BOLD-LLM |
|---|---|---|---|---|
| (a) Reduction of time and effort spent | No | Maybe | Yes | Yes |
| (b) Controlling the search space in the relevant problem and solution domains | Maybe | Maybe | Maybe | Yes |
| (c) Automation | No | No | Yes | Yes |
| (d) Making inferences about solution approaches utilized (relevancy, popularity, trend) | No | No | No | Yes |
| (e) Pros and cons | Maybe | Yes | Maybe | No |

The work of H. Chen et al. [62] produce a written section of relevant background material to a solution approach written in the form of a research paper through a BERT-based semantic classification model. Similarly, K. Heffernan et al. [63] utilize a series of machine learning algorithms as automatic classifiers to identify solutions and problems from non-solutions and non-solutions in scientific sentences with good results. These findings suggest that ML-based language models can be utilized in the automation of literature review with success.

Consequently, literature that explains the procedure of manually and automatically reviewing the literature is determined. Also, automated tuning frameworks for different modeling schemes are identified.

In Table 2, the benefit and function features of the methods used in the solution approach selection are compared. Features include saving time and effort (a), ensuring that the required problem and solution domains are searched for using a systematic keyword selection scheme (b), automating tasks (c), drawing conclusions about the relevance, popularity, and trend of the solution approaches used for the use case (d), and the pros and cons information of selected methods (e). The effectiveness of the methods used in a research paper for a specific use case is determined by relevancy metrics, which indicate how well the methods align with the specificity of the use case. The popularity metric is used to assess the research interest of a paper and the methods used in the paper. It is calculated by considering the number of citations and the age of the publication. On the other hand, trend analysis based on a total number of papers that use a specific solution approach provides insights, making knowledgeable decisions, and planning by examining trends and behaviors throughout time.

It is seen that there is a gap in the methods of solution approach selection in terms of satisfying the specified features. This article aims to investigate and address this gap to obtain a tool with all the features listed in Table 2.

## 3 Methodology

SARBOLD-LLM has three main modules illustrated by the flowchart in Figure 2. They are called "Module 1: Scopus search", "Module 2: Scoring and method extraction", and "Module 3: Analyzes", respectively. Red ellipses are the start and end points, green parallelograms are inputs and outputs, blue rectangles are processes, orange diamonds are decisions, and the purple cylinder is the database. The red dashed line demonstrates the automated flow. The first module named "Scopus search" covers selecting keywords and getting results via Scopus Search.[2] Then the advanced search query returns the results where the fields are explained by Scopus Search Views.[3] In the second module named "scoring and method extraction", solution methods that are used in each article are searched using the OpenAI API. In the third module named "analyzes", sensitivity and post-analyzes are performed. The flow indicated by the red dashed line is performed automatically.

The current version of SARBOLD-LLM is given in Algorithms 1 and 2. It takes as input the table of keywords used to build the Scopus query (It could also take a query as input to allow more liberty after some iteration) and the prompt used for the OpenAI API. It provides at the end a list of methods, with associated scores (number of papers, citations, relevancy, which can be viewed year by year to detect trends) and papers. To help the end user to select and filter methods, an automated clustering has been made. It allows filtering methods used only one time and regroups similar names such as YOLO-v3 and YOLO-v4.

---

[2] https://dev.elsevier.com/sc_search_tips.html
[3] https://dev.elsevier.com/sc_search_views.html

**Algorithm 1**: SARBOLD-LLM

> *Input:* a list of lists of keywords (m_keywords) and a prompt (m_prompt)
> *Output:* A list of methods with scores and associated papers
> 1    scopus_query ← **make_query**(m_keywords)
> 2    unsatisfied ← True
> 3    *While* (unsatisfied) *do*
> 4       paper_list_data ← **scopus_api**(scopus_query)
> 5       unsatisfied ← **ask_user**(paper_list_data)
> 6       *If* (unsatisfied) *do*
> 7          (paper_list_data, unsatisfied) ← **manual_cleaning**(paper_list_data)
> 8          *If* (unsatisfied) *do*
> 9             m_keywords ← **modify_keywords**()
> 10            scopus_query ← **make_query**(m_keywords)
> 11   paper_scores ← **compute_paper_scores**(paper_list_data, m_keywords)
> 12   paper_methods ← **openai_api**(paper_list_data, m_prompt)
> 13   clustered_methods_with_data ← **clustering**(paper_methods, paper_scores)
> 14   *Return* clustered_methods_with_data

**Algorithm 2**: clustering

> *Input:* a list of methods with associated data (paper_methods) and a list of scores (paper_scores)
> *Output:* A list of clustered methods with scores and associated papers
> 1    method_names ← **extract_names**(paper_methods)
> 2    metric ← 1 - **normalized_indel_similarity**(methods_names, methods_names)
> 3    clusters ← **dbscan**(method_names, metric, eps←0.2, min_samples←2)
> 4    clustered_methods_with_data ← []
> 5    *For* i *in* clusters *do*
> 6       data_and_scores ← []
> 7       *For* j *in* method_names *do*
> 8          *If* j *belongs to* i *do*
> 9             data_and_scores ← **add_data**(**retrieve_paper_and_score**(j))
> 10      clustered_methods_with_data ← **add_cluster**(i, data_and_scores)
> 11   clustered_methods_with_data ← **clean_false_pos**(clustered_methods_with_data)
> 12   *Return* clustered_methods_with_data

SARBOLD-LLM scheme is appropriate for any problem and solution domain. It can be used for use cases in many different fields. Although the second module of this study focuses on AI methods, this module can also evolve into other topics, such as which hardware to be used and which scientific applications to be employed. However, as the SARBOLD-LLM relies on the OpenAI framework, ground truth data is created manually (by writing AI methods from the title and abstract of papers) to check the performance.

### 3.1 Module 1: Scopus Search

The goal of the first module is to search for a relevant pool of papers concerning the given problem a user is dealing with. To do so, a keyword selection scheme has been made to facilitate the user's work. This scheme is then used to make a Scopus query, but also to score each paper.

To determine keywords, three specification levels (a general, an expanded, and a detailed one) are applied to the given problem and the searched solutions. This work is done manually as it involves eliciting user information on the use case. That means both classification and order are specified by the user. However, this stage is critical in recommending more appropriate solution approaches because these keywords are the first inputs to the SARBOLD-LLM and determine the pool of papers used in module 2 anointed "scoring and method extraction". The different levels showcase;

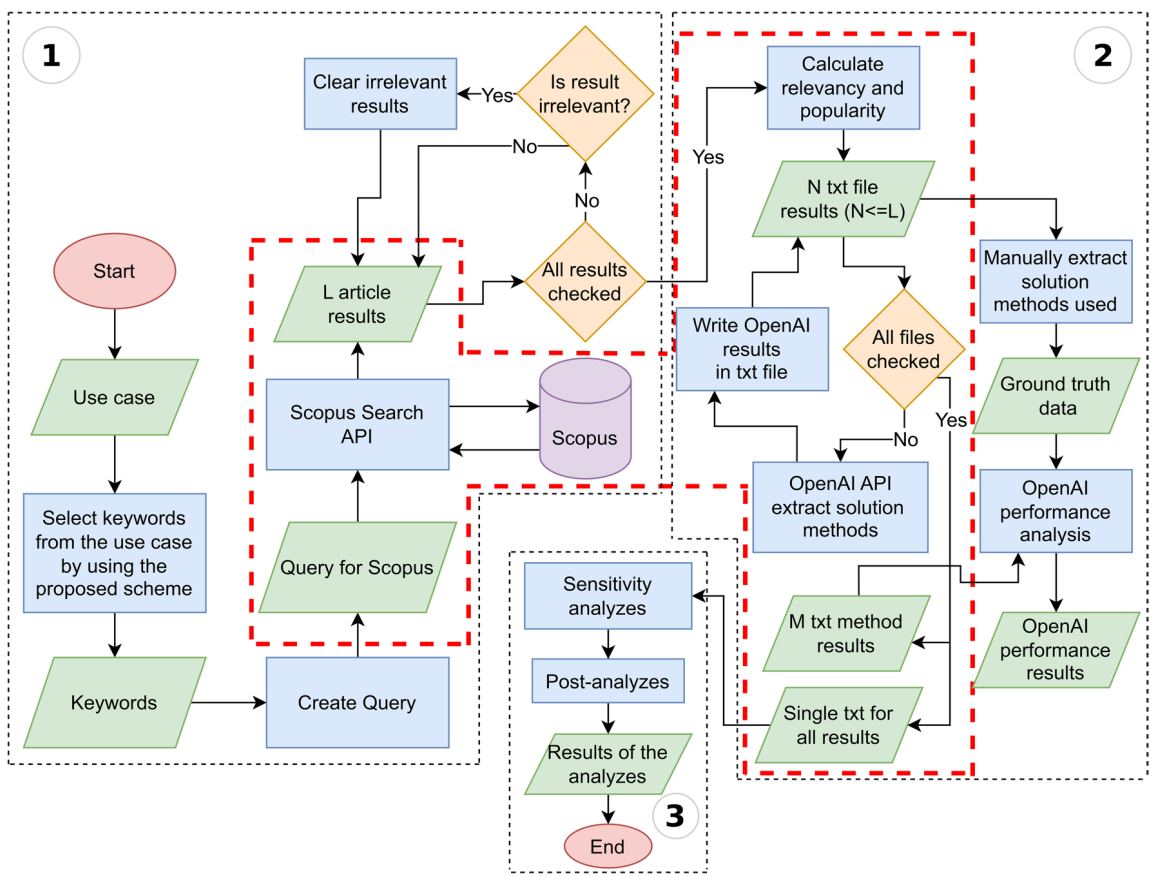

**Fig. 2** The flowchart of the SARBOLD-LLM

Level 1 The general and necessary keywords. The keyword must be a part of the research paper for the paper to be in the selected pool of papers.
Level 2 The expanding keywords. Here only one of the keywords in the field is necessary for the paper to be selected.
Level 3 A further specification, use case-specific keywords. It is only used in the later stage to rank the identified solution methods with the relevancy metric.

Figure 3 gives an example of the proposed keyword selection scheme. The problem domain block covers the specific area or subject matter that a problem or project deals with. The scope and context of the issues that need to be addressed are defined by the problem domain. On the other hand, the solution approach, also known as the solution space, is the strategy or method used to address and solve the problems identified within the problem domain. It outlines how you plan to design, implement, and deliver a solution to the issues at hand.

These two blocks contain keywords according to the levels explained above. In the database search, adding some version of a keyword will only search for that specific keyword. Consequently, other versions of that word will also have to be added, but with the logic operator OR to indicate that either version can be used in the paper, for instance, "AI OR artificial intelligence".

Notice that it is possible, but not necessary, to add keywords in each field, where a field refers to the specific level in the block. Leaving some fields empty will lead to a less specified pool of solution approaches, which consequently risks not fitting the use case. At the same time, adding too many keywords can lead either to a too restricted pool of papers (e.g., if one uses too many general keywords, and fulfills each field) or, if too many expanding keywords are given, to a less specific pool of paper as if the field was left empty.

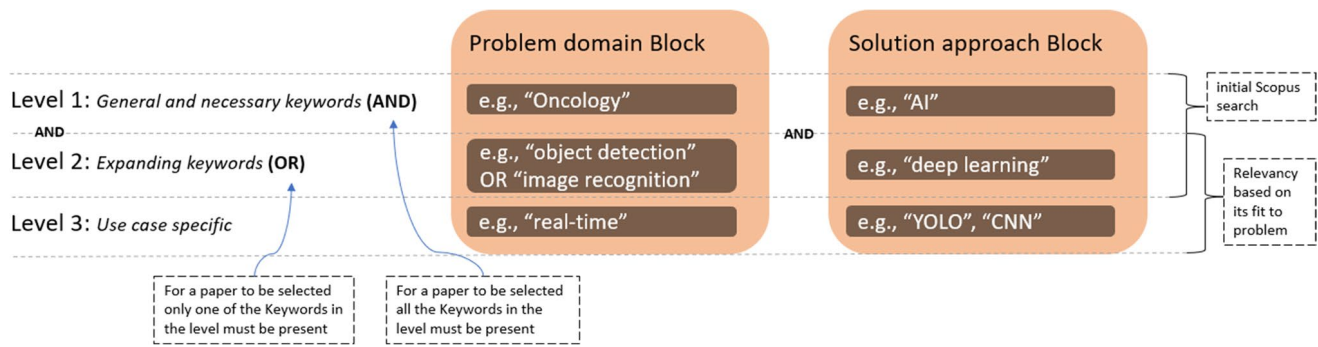

**Fig. 3** Example illustration of the proposed framework for keyword selection

After keyword selection, a query is created for Scopus Search API. Information is searched in titles, abstracts, and keywords of recent articles or conference papers, for the words defined in levels 1 and 2. The query can be, for example:

'TITLE-ABS-KEY(("oncology") AND ("artificial intelligence" OR "AI") AND ("image processing")) AND DOCTYPE(ar OR cp) AND PUBYEAR > 2013'

Note that an expert can directly enter a query instead of using the keyword selection scheme. It is useful in some cases, for example: when it is difficult to find a good pool of papers using the query built by the keyword selection scheme, or when one wants to search in a specific field or a specific range of years, or for a first try if one wants to search only for reviews to get more appropriate academic keywords. However, it is still advantageous to follow this scheme as it helps to find, classify, and order the use case keywords, but also to specify what is important for scoring the paper.

Another way to help the end user get the query could be to extract keywords directly from a summary of the given use case. However, it is difficult to automate a keyword extraction scheme for several reasons. First, one needs to distinguish keywords used for establishing the search space and keywords used for scoring papers (or methods). Secondly, the query is sensitive to each keyword and it is often necessary to change the list of keywords used to get a more appropriate pool of papers, using synonyms or re-ordering some keywords. The third reason concerns the end user itself, as this scheme helps the end user to understand what are the most important features/constraints of his use case, and what is recommended but not necessary.

The publication year, the number of citations, the title, and the abstract information of all articles returned by the Scopus query are saved. After all the results are obtained, the title and abstract information of all the articles are examined manually, and articles that are irrelevant and have not applied/mentioned any AI method are eliminated.

### 3.2 Module 2: Scoring and Method Extraction

In this module, the relevancy and popularity metrics for the Scopus search results are computed, and solution methods are extracted from the title and abstract of each paper.

The relevancy metrics count the number of unique level 2 and 3 keywords appearing at least once in the title, abstract, or keywords. Ultimately, the metric represents how well the methods fit the specificity of the use case. For example, a paper named "Hybrid learning method for melanoma detection" yields in the abstract "image recognition (5 times), deep learning (2 times), real-time"; it will therefore have a relevancy metric of 3, taking into account Fig. 3.

The popularity metric is used to know the research interest of a paper and its methods. It is computed by

$$\frac{citation\ number}{publication\ age\ in\ whole\ years + 1},$$

where 1 is added in the denominator to avoid zero divisions, and the citation number is obtained from the Scopus database.

After calculating the relevancy and popularity metrics, the SARBOLD-LLM inputs the title and abstract information to OpenAI and outputs the AI approaches used in each article.

When someone provides a text prompt in OpenAI API, the model will produce a text completion that tries to match the context or pattern you provided. Essential generative pre-trained transformers (GPT)-3 models, which generate natural language, are Davinci, Curie, Babbage, and Ada. In this article, "text-davinci-003" is used which is the most potent GPT-3 model and one of the models that are referred to as "GPT 3.5".[4] Some issues to consider when preparing prompts are as follows[5]:

---
[4] https://beta.openai.com/docs/model-index-for-researchers
[5] https://help.openai.com/en/articles/6654000-best-practices-for-promptengineering-with-openai-api

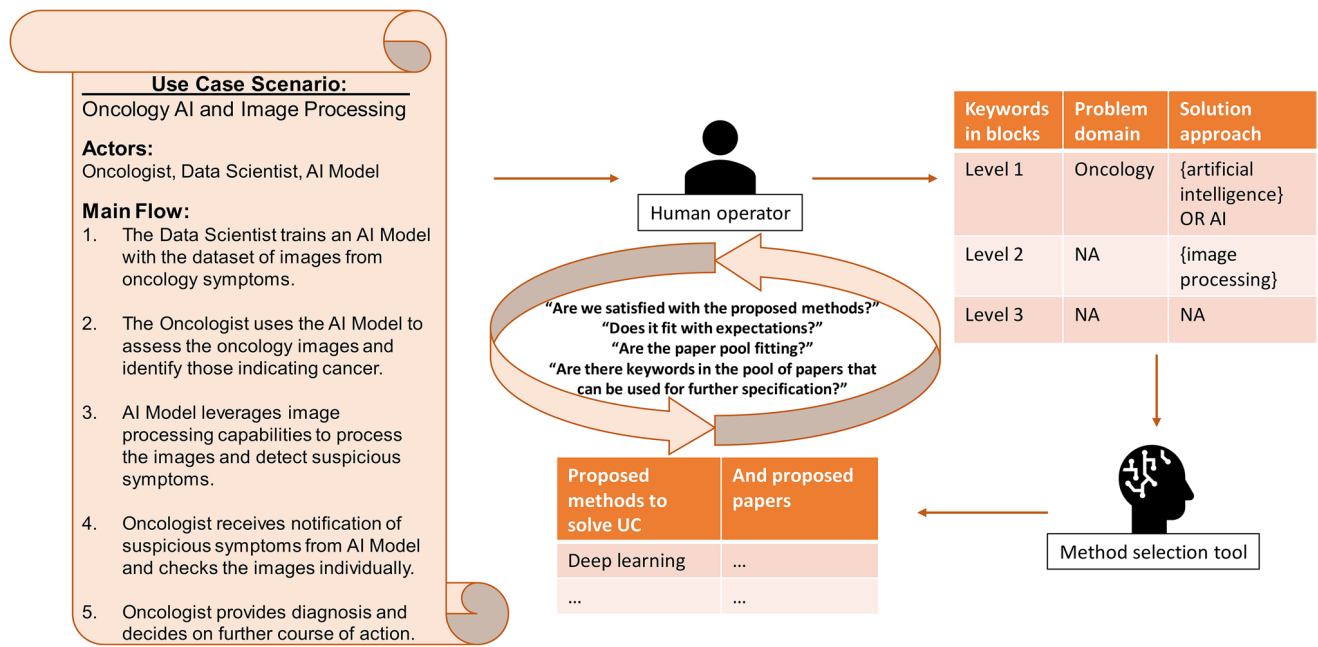

**Fig. 4** An example use case and an illustration of the procedure of interactively using the SARBOLD-LLM

- It is advised to place instructions at the start of the prompt and to use ### or"""" to demarcate the context from the instruction.
- Speaking of what to do is preferable to speaking about what not to do.

The prompt can then be the following:

"Extract the names of the artificial intelligence approaches used from the following text. ###{" + str(document_text) + "}### \nA:"

where 'document_text' includes the title and abstract information of a paper.

To evaluate OpenAI's performance, the ground truth AI methods are manually produced for non-filtered papers, regarding each paper's title and abstract information. Some high-level tags, such as "artificial intelligence" and "machine learning" are not included. In other words, the keywords used in Scopus search as a method are not involved. Precision, recall, and F1-measure are calculated for performance analysis.

### 3.3 Module 3: Analyzes

Firstly, sensitivity analyzes are done regarding Scopus and OpenAI in this module. Different combinations of level 1 and 2 keywords in the Scopus query are tried and the initial prompt is compared with other prompts for OpenAI.

For the selected use case, post-analyzes are performed by investigating which AI methods are used more often and which have higher relevancy or popularity metrics and comparing the results over different periods. This can be done manually, or, if there are too many methods listed, first a clustering algorithm can be used to help this investigation. Currently, DBSCAN [64] is used with (1 − the normalized Indel similarity) as distance performs well enough to support post-analysis. In the controlled comparison, it is seen that the clusters made manually and with DBSCAN are very close to each other for the same observed data.

## 4 Experiments

### 4.1 Use Case Definition

The use case example given in Figure 4 is tackled for the initial experiment. Here, AI is employed on the dataset of images to detect cancer.

### 4.2 Keywords From the Use Case Scenario

Using Fig. 3, the following keywords are defined: "oncology" as problem level 1, "artificial intelligence" and "AI" as solution level 1. Only "image processing" is used as solution level 2. By using only one level 2 keyword, the experiment stays rather general in the expected results.

For simplicity, level 3 keywords are not used in this example. Level 3 keywords do not affect the pool of papers but enable the user to elicit relevancy to papers that match their use case better. Because the computation of the relevancy metric is trivial, it is omitted in this example.

## 4.3 Scopus API Search and Manual Article Cleaning

According to the selected keywords, our initial query of Scopus API[6] is given below.

'TITLE-ABS-KEY(("oncology") AND ("artificial intelligence" OR "AI") AND ("image processing")) AND DOCTYPE(ar OR cp) AND PUBYEAR > 2013'

That means the keywords are searched in the title, abstract, and keyword parts. In addition, to limit the size of the results, the publications published after 2013 are selected, and to be more specific, the document type is restricted to "Article" or "Conference Paper".

Digital object identifier (DOI), electronic identifier (EID), year, and citation number results that Scopus API returns are given in Table 6. The relevancy and popularity values are calculated as stated in Section 3.2. Currently, some papers can have a relevancy of 0, but by manually checking them, they stay relevant. It happens when keywords only appear in "INDEXTERMS" provided by Scopus but are absent from the title, abstract, and author keywords. Moreover, this is also due to a total absence of keyword level 3. It can be fixed by taking these automatic keywords for the OpenAI analysis.

The query returns 92 results. Among them, 25 publications (irrelevant, not technical, just survey, etc.) indicated in red in Table 6 are manually filtered. The remaining 67 articles are the results related to the domains and keywords of the use case. However, there are among them 12 papers, highlighted in orange, that apply an AI method successfully, but they do not mention particular methods (they do only highly general, level 1 and 2 ones) in the title and abstract; they will therefore be missed by the OpenAI extraction part that is stated in Section 4.4. However, it is not critical as trends are explored. Still, 55 papers remain to be analyzed. Note that of the 37 articles eliminated, these could have been marked as such if we had implemented the level 3 keywords.

## 4.4 OpenAI

The initial prompt for the OpenAI API is stated below.

"Extract the names of the artificial intelligence approaches used from the following text. ###{" + str(document_text) + "}### \nA:"

where 'document_text' includes the title and abstract information of a paper.

After finding methods using OpenAI and manual work, the performance values are calculated. It is assumed that manual findings are the actual methods. On the other hand, the results coming from OpenAI are the predicted ones.

### 4.4.1 OpenAI Performance

To analyze the results, the methods found by OpenAI are compared to the ones found by manual investigation (considered ground truths) for each paper. There are different performance determinants:

- "*true found*" is the number of methods found by OpenAI that belong to the ground truths,
- "*false found*" is the number of methods found by OpenAI that do not belong to the ground truths,
- "*true general found*" is the number of methods found by OpenAI and the manual search but belonging to level 1 or 2 keywords or high-level keywords like "machine learning",
- "*total manual*" is the number of ground truths,
- "*missing*" = "*total manual*"—"*true found*".

With these data, precision, recall (or sensitivity or true positive rate), and F1-score can be calculated for performance analysis. To do that, the following metrics are employed:

- True Positive (TP) = "*true found*",
- False Positive (FP) = "*false found*" + "*true general found*",
- False Negative (FN) = "*missing*".

The "*true general found*" results are counted as FP since they are terms that are entered into the Scopus search or they are high-level keywords for our solution domain interest like "machine learning, artificial intelligence-based approach" as mentioned above.

For each paper that is not filtered, the performance metrics are calculated as follows.

$Precision = TP/(TP + FP)$
$Recall = TP/(TP + FN)$
$F1 - score = \frac{Precision x Recall}{(Precision+Recall)}$

The F1-score assesses the trade-off between precision and recall [65]. When the F1-score is high, it indicates that both precision and recall are high. A lower F1-score indicates a larger imbalance in precision and recall.

Let's check the following example, coming from [66]: "Transfer learning with different modified convolutional neural network models for classifying digital mammograms utilizing Local Dataset".

---

[6] https://dev.elsevier.com/sc_search_tips.html

" [...] accuracy of different machine learning algorithms in diagnostic mammograms [...] Image processing included filtering, contrast limited adaptive histogram equalization (CLAHE), then [...] Data augmentation was also applied [...] Transfer learning of many models trained on the Imagenet dataset was used with fine-tuning. [...] NASNetLarge model achieved the highest accuracy [...] The least performance was achieved using DenseNet169 and InceptionResNetV2. [...]"

Manually, "transfer learning", "convolutional neural network", "NASNetLarge", "DenseNet169","InceptionResNetV2", "data augmentation", and "fine-tuning" are found as AI methods. What OpenAI has found is indicated as well. Firstly, "transfer learning", "convolutional neural network", "data augmentation", 'NASNetLarge", "DenseNet169" and "InceptionResNetV2″ are *"true found"*; so $TP = 6$. Secondly, "machine learning algorithms" is a *"true general found"*, and "contrast limited adaptive histogram equalization (CLAHE)" is "false found", then $FP = 2$. Finally, "fine-tuning" is a *"missing"* and so $FN = 1$. With these, one can compute $Precision = 6/(6+2) = 0.75$, $Recall = 6/(6 + 1) = 0.86$ and $F1\text{-}score = (2 \times 0.75 \times 0.86)/(0.75 + 0.86) = 0.8$.

In our studied case (see Appendix 2), the average scores are good, with an average precision of 0.7111, recall of 0.9226, and F1-score of 0.7775. There are 108 TPs, 51 FPs, and 12 FNs if all 55 results are grouped into a single result pool. Then the values of the precision, recall, and F1-score are 0.6793, 0.9, and 0.7742, respectively. All ground truths and OpenAI findings are presented in Table 7.

A manual literature review takes a week to complete, but the SARBOLD-LLM completes the whole task in a few hours (selecting AI approaches from the title and abstract of unfiltered 92 publications).

## 4.5 Sensitivity Analyzes

### 4.5.1 Scopus API Sensitivity

For the Scopus sensitivity analysis, different combinations of level 1 keywords are tried in the query. The initial query can be seen in Section 4.3.

Table 3 shows the impact of changing keywords in level 1. The first query in Table 3 is the initial one, given for comparison. Changing a problem domain keyword with another that could be seen as a synonym can greatly impact the papers found. Using the more specific keyword "machine learning" in the solution domain instead of "artificial intelligence" has an impact on the publications found. Similarly, in the problem domain using "cancer" instead of "oncology" has a great impact on the number of papers found. On the other hand, changing double quotes to braces does not have that much effect. Moreover, it seems that using only an abbreviation instead of an open form can change the number of results found. Using only the abbreviation has resulted in a poor paper pool.

**Table 3** Summary table for different queries

| Query level 1 keywords problem/solution | Papers found | Common papers with the initial query |
|---|---|---|
| "oncology"/ "artificial intelligence" "AI" | 92 | 92 |
| "cancer"/ "artificial intelligence" "AI" | 746 | 64 |
| "oncology"/ "machine learning" "ML" | 155 | 53 |
| {oncology}/ {artificial intelligence} {AI} | 92 | 92 |
| "oncology"/ "artificial intelligence" | 89 | 89 |
| "oncology"/ "AI" | 16 | 16 |

However, despite the different pool of papers, the methods found by OpenAI are pretty much the same, both for the second and the third query. This means that using synonyms changes the pool of papers but not the methods used to solve the same kind of problem, which means that the method is robust to the keyword selection scheme.

### 4.5.2 OpenAI Sensitivity

To analyze the sensitivity of OpenAI, different prompts are tested, and the differences between proposed AI methods are checked. Results are summarized in Table 4, and details are provided in Table 8, Table 9, and Table 10 in Appendix 3. The below prompts are used for analysis.

**Table 4** Summary table for OpenAI sensitivity concerning the initial prompt

|  | Total number of missing / extra or different words | Total number of articles which have the same results | Column 2 divided by column 1 |
|---|---|---|---|
| Prompt 1 | 117 | 12 | 0.1026 |
| Prompt 2 | 107 | 16 | 0.1495 |
| Prompt 3 | 101 | 15 | 0.1485 |
| Prompt 4 | 31 | 41 | 1.3226 |
| Prompt 5 | 96 | 18 | 0.1875 |
| Prompt 6 | 51 | 34 | 0.6667 |

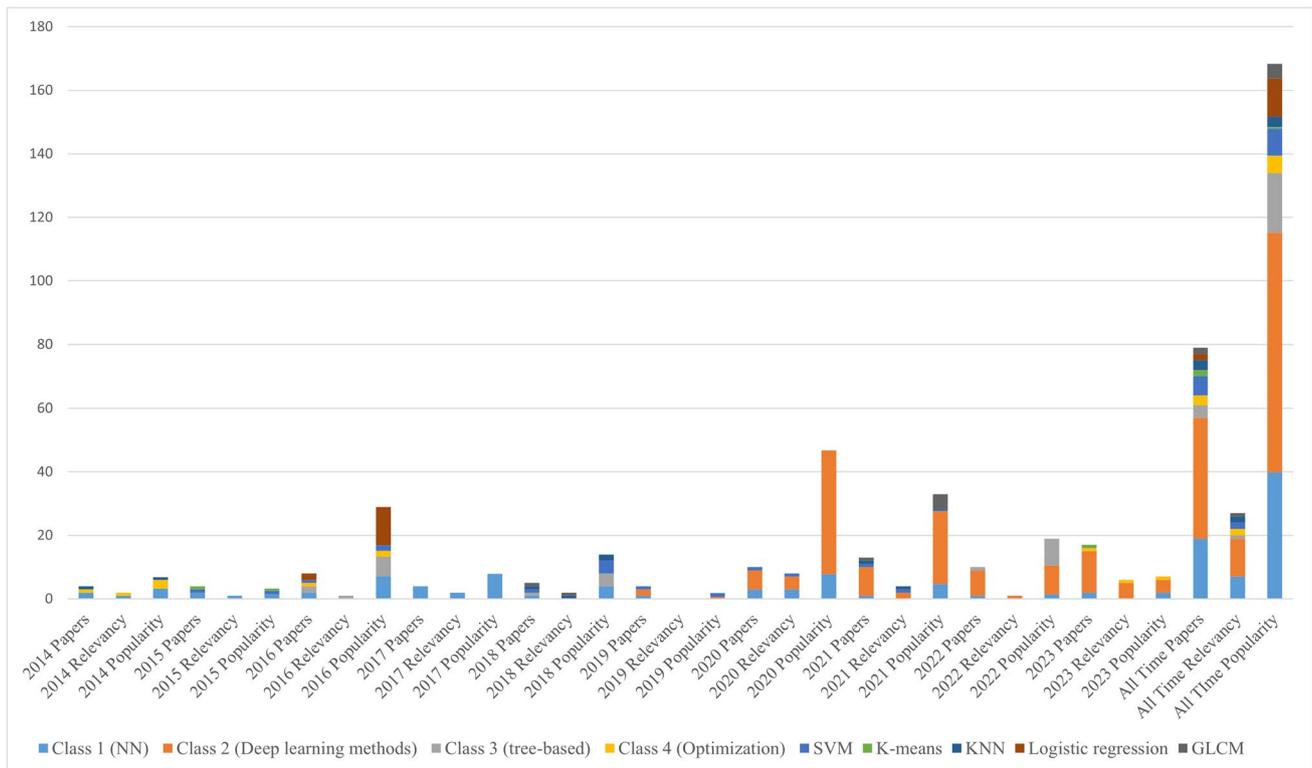

**Fig. 5** Summary chart of extracted AI methods for "oncology" problem domain and "image processing" solution approach

"Extract the names of the artificial intelligence approaches used from the following text. ###{" + str(document text) + "}### \nA:"

Prompt 1

"**Just write** the names of **used** artificial intelligence **or machine learning methods in** the following text. ###{" + str(document text) + "}### \nA:"

Prompt 2

"**Just write** the names of **used** artificial intelligence **methods in** the following text. ###{" + str(document text) + "}### \nA:"

Prompt 3

"**Just write** the names of artificial intelligence approaches used **in** the following text. ###{" + str(document text) + "}### \nA:"

Prompt 4

"Extract ~~the~~ names of the **used** artificial intelligence approaches from the following text. ###{" + str(document text) + "}### \nA:"

Prompt 5

"**Write** the names of **successfully applied** artificial intelligence approaches **in** the following text. ###{" + str(document text) + "}### \nA:"

Prompt 6

"Extract the names of ~~the~~ artificial intelligence approaches **employed in** the following text. ###{" + str(document text) + "}### \nA:"

In Table 4, the number in the last column is an enriched ratio, meaning that if two prompts are equal, it will obtain an infinite value. However, having a difference between two prompts will lead to a decreasing ratio, considering that two papers do not provide the same set of words but also how many words in the prompt are different.

The original prompt has a higher F1-score value than the other six prompts. With these few prompts, it can already be said that OpenAI is sensitive to the sentence used. However, it generally adds words with respect to the manual search, and extracting the most common words belonging to these results should be enough to find what the user is searching for. Moreover, it is observed that changing a word's position has less impact than changing a word; the more words the user changes, the more differences appear. It also seems that using more common/usual words will give more generic results, closer to the ones that are being searched for; when using very specific instructions, notably in the action verbs, the results will generally be more irrelevant.

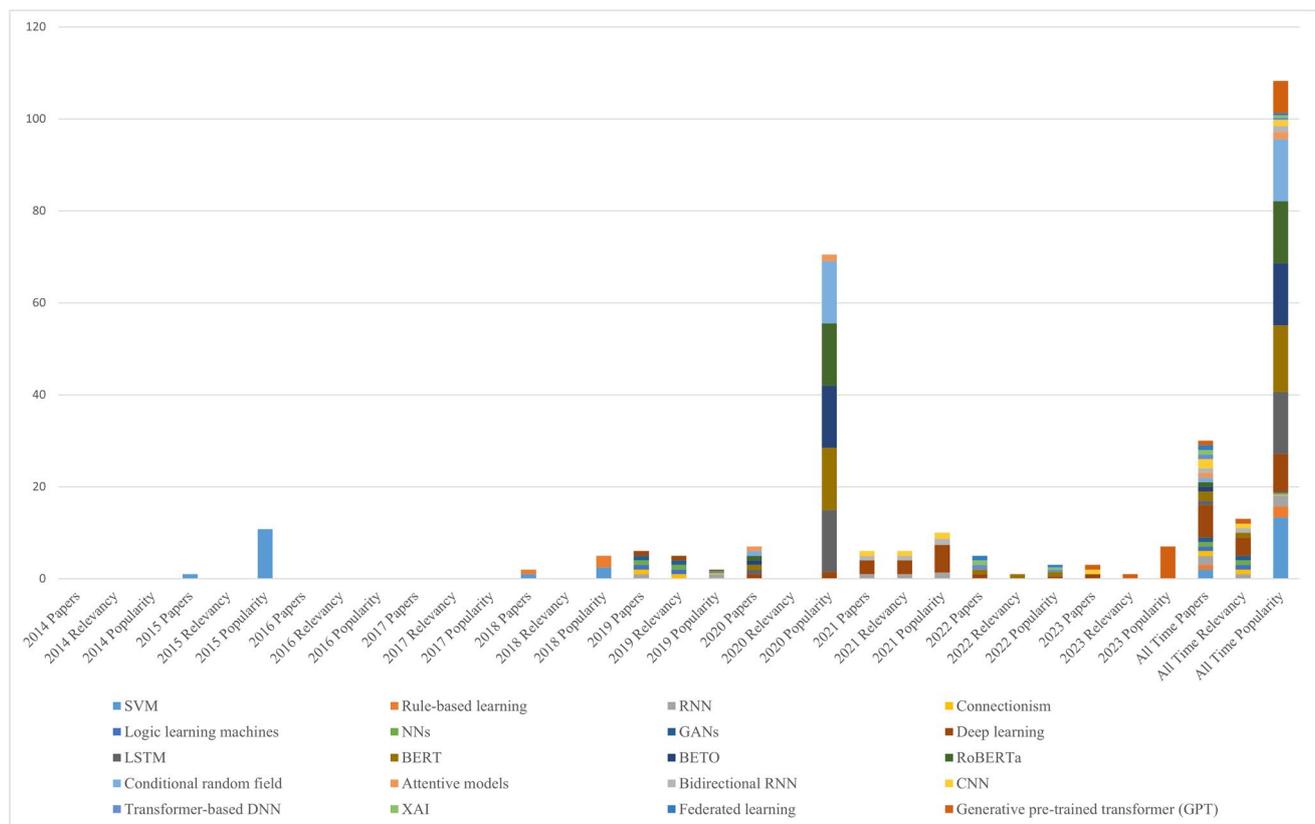

**Fig. 6** Summary chart of extracted AI methods for "oncology" problem domain and "natural language processing" solution approach

### 4.6 Post-Analyzes

The extracted AI methods for the use case described in Section 4.1 are presented in Appendix 4. The total number of appearances of the methods, their relevancy, and popularity metrics are showcased in Table 11 by years. Methods selected from articles that are not highlighted in Table 6 and appeared at least in two papers are discussed.

Figure 5 illustrates the summary chart of Table 11. It is seen from the figure that many different methods have been investigated to solve our example use case, but some are much more used or popular than others. These methods (e.g., class 2 (deep learning methods) and class 1 (artificial neural networks)) are the ones that the user should investigate in the first place to solve the given use case. To be more specific, until 2018 different types of neural networks, logistic regression, SVM, and random forest are popular methods. After 2018, SVM and neural networks are still utilized, and the extra trees classifier seems popular in 2022. However, the trend is being dominated by deep learning methods. Among the deep learning algorithms, CNN, U-Net, and AlexNet can be counted as the three most used and popular methods.

AI methods can be examined without making any classification, but in this case, there will be too many methods. To simplify this situation, the methods are divided into classes. In Appendix 4, specifics on method classification and detailed information for AI methods in these classes are provided. Moreover, a more detailed decision-making process can be made by using relevancy and popularity metrics. For example, these metrics support decision-making when being uncertain between two AI methods.

### 4.7 Experiments for Different Problem Domains

To check the robustness of the SARBOLD-LLM, different problem domains, and solution approaches are also considered for the Scopus search. The same initial prompt given in Section 4.4 is used for all use cases to extract AI methods by utilizing OpenAI API.

First, the same problem domain is kept, and the level 2 solution approach is changed as given in the below query.

'TITLE-ABS-KEY(("oncology") AND ("artificial intelligence" OR "AI") AND ("natural language processing" OR "NLP")) AND DOCTYPE(ar OR cp) AND PUBYEAR > 2013'

The aforementioned search yields 35 documents. Although 5 of them effectively use an AI approach, they do not mention any particular methods in the title or abstract, and 15 of them are irrelevant or merely surveys. Consequently, 15 of them are selected in the manner described in Section 4.3. Figure 6 shows AI methods employed in selected papers. Until 2019, SVM seemed to be a popular method, and from 2019 the trend has shifted to deep learning algorithms. RNN, CNN, and BERT are among the deep learning methods that are more used after 2019. In addition, some of the most popular methods are BERT, LSTM, and GPT.

Secondly, the solution approach components are retained the same while changing the problem domain. The query for the "traffic control" issue domain is presented below.

'TITLE-ABS-KEY(("traffic control") AND ("artificial intelligence" OR "AI") AND ("image processing")) AND DOCTYPE(ar OR cp) AND PUBYEAR > 2013'

The query returns 52 results, where nine are irrelevant or just surveys, and 20 use an AI method successfully, but they do not mention specific methods in the title and abstract. Therefore, 23 of them are selected. In Fig. 7, it is seen that until 2020, classical methods like SIFT, SURF, KNN, and decision trees are popular methods. After 2020, the deep learning methods class (that contains R-CNN, fast R-CNN, faster R-CNN, YOLO, deep simple online real-time tracking (DeepSORT), CNN, U-Net, etc.) is on the rise in terms of the number of uses and popularity.

Another query is the "satellite imagery" for the problem domain, given below. It returns 66 results and 37 of them are selected to be used in analyzes.

'TITLE-ABS-KEY(("satellite imagery") AND ("artificial intelligence" OR "AI") AND ("image processing")) AND DOCTYPE(ar OR cp) AND PUBYEAR > 2013'

Figure 8 illustrates the summary of extracted AI methods for the "satellite imagery" problem domain. Class 1 includes CNN, DNN, DeepLabv3 +, FCN, U-Net, U-Net + +, encoder-decoder, attention mechanism, Res2Net, ResNet, LSTM, SegNet, V-Net, U2Net, AttuNet, LinkNet, mask RCNN, and cloud attention intelligent network (CAI-Net). On the other hand, class 2 covers ant colony optimization (ACO), genetic algorithm, particle swarm optimization (PSO), bat algorithm, and artificial bee colony (ABC). Until 2020, SVM, ANN, and ACO were frequently used and popular methods. After 2020, the use and popularity of class 1 and PSO appear to be increasing. In class 1, the top three most used and most popular methods are CNN, U-Net, and DNN. As can be seen from the trend, the first methods to be considered in this problem domain may be the deep learning methods given above.

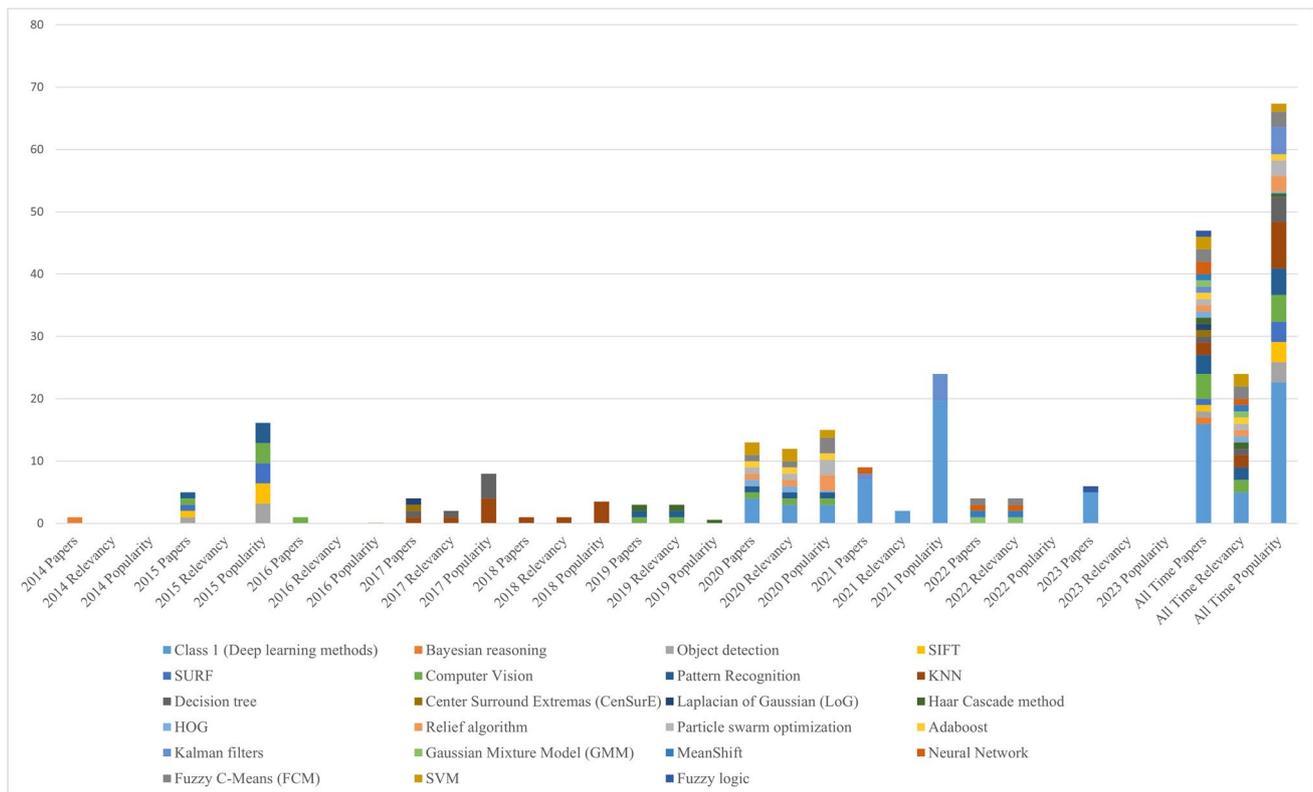

**Fig. 7** Summary chart of extracted AI methods for "traffic control" problem domain and "image processing" solution approach

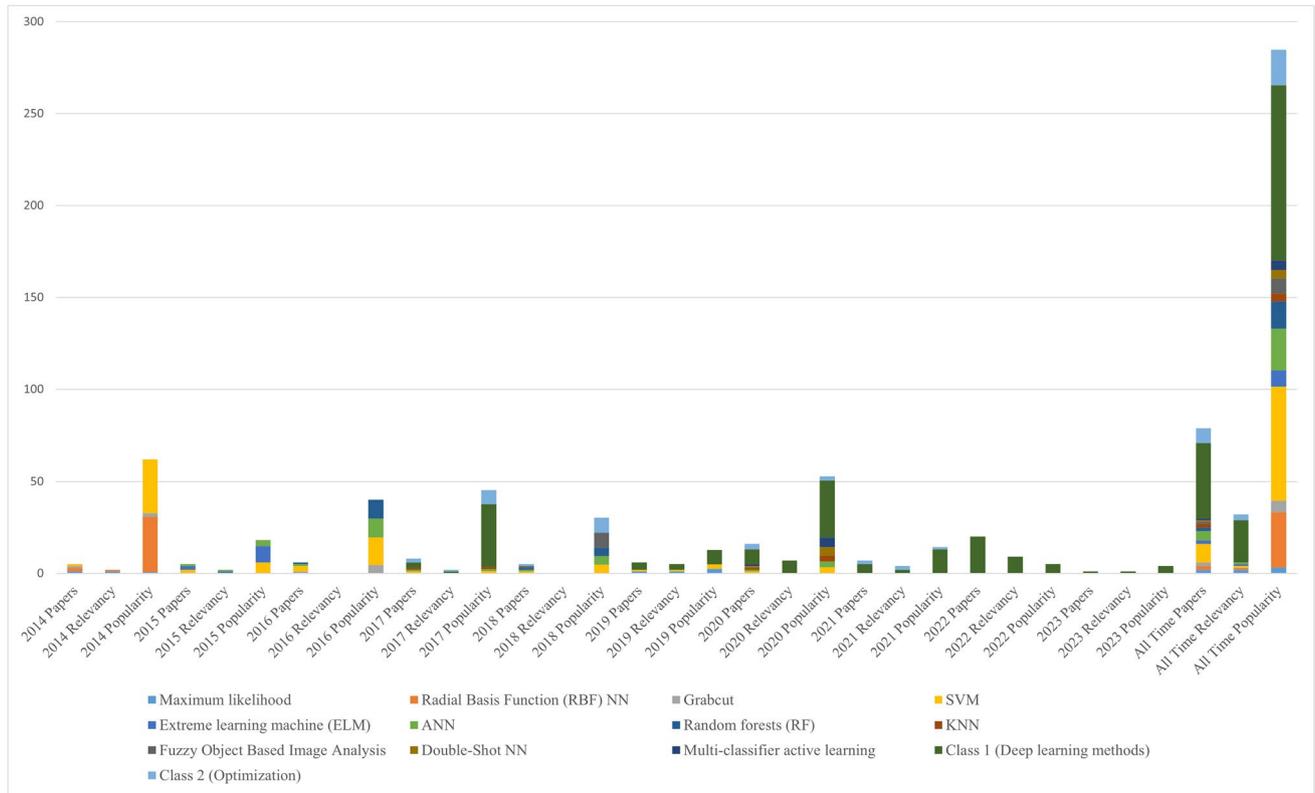

**Fig. 8** Summary chart of extracted AI methods for "satellite imagery" problem domain and "image processing" solution approach

In Table 5, OpenAI performance results for all experiments are given, where TP, FP, and FN values are considered as a single pool, i.e., performance metrics are not average values for each article result. It should also be taken into account that if the *"true general found"* words (i.e., machine learning, artificial intelligence, image processing) are not included in the FP, higher precision and F1-score values would have been obtained. Although the problem domain and solution approach change, similar performance results are attained, which is promising for the robustness of the SARBOLD-LLM.

## 5 Discussion

A big issue when utilizing automatic solution method selection schemes is the trust in the fit, relevancy, and popularity of the suggested methods. The fit to the actual use case depends on the ability of the human operator to interact with the tool and whether or not they understand the intricacies of the approach. With the SARBOLD-LLM tool, the human operator has the ability to validate the suggested methods from the accompanying pool of research papers, and due to the simplicity, responsiveness, and intuitiveness, it is relatively straightforward for the human operator to modify and align the usage of the tool with the overall goal of solving a problem. Additionally, to increase the tool's performance in terms of operation requirements (e.g., explainability, trustworthiness) and resources (e.g., hardware), the necessary features or extra resources for AI methods can be added and expanded later if the detailed requirements and current resources are stated clearly.

**Table 5** OpenAI performance results for all problem domains

| Problem domain | Solution approach | Precision | Recall | F1-score |
|---|---|---|---|---|
| oncology | (artificial intelligence ∨ AI) ∧ image processing | 0.6793 | 0.9 | 0.7742 |
| oncology | (artificial intelligence ∨ AI) ∧ (natural language processing ∨ NLP) | 0.6667 | 0.9677 | 0.7895 |
| traffic control | (artificial intelligence ∨ AI) ∧ image processing | 0.6026 | 0.9216 | 0.7287 |
| satellite imagery | (artificial intelligence ∨ AI) ∧ image processing | 0.7917 | 0.9406 | 0.8597 |

For example, if explainability is required, many different methods exist for obtaining explainable AI (XAI) methods [67–72]. On the other hand, if trustworthiness is required, then according to the system, environment, goals, and other parameters where AI will be used, several alternative criteria for trustworthiness may be specified [73, 74].

Details or requirements such as explainability and trustworthiness can be retrieved in the keyword selection scheme in Fig. 3. Or, after AI methods are found by the SARBOLD-LLM, post hoc analyzes can be made with the requirements not used in the SARBOLD-LLM. In some use cases, such requirements or details may not be specified at the beginning of the AI system life cycle and, therefore, may not be included in the keyword selection phase.

Due to the specificity of certain use cases, there is a considerable risk that no research has been conducted on the specifics of the use case. Consequently, the proposed solution approach methods will likely not showcase a high score in the relevancy metric. Therefore, the literature pool must be investigated after the results are identified.

Ultimately, the SARBOLD-LLM's applicability comes down to the objective of the application. It will comfortably propose methods already explained in the literature as to why it is very useful when identifying trends in the research communities. However, as the method identification is based on historical data that train the tool to determine what words within a research paper can be classified as a method, the tool will not fare well when dealing with entirely new solution approach schemes.

It is noteworthy that the relevancy explained in Section 3.2 is computed and saved at the same time as the other data. It could be useful in the future if one wants an automatic filter. On the other hand, if the pool of papers is too big to be manually filtered, it is possible to filter at the end of the process, when one is checking for the methods to be used. The main disadvantage of filtering after the whole process is that it can allow a lot of irrelevant papers to be analyzed by OpenAI, and this will modify the perception of the trends of research for the studied use case. However, note that SARBOLD-LLM is used to get trends in research about a given use case to support the selection of solution methods, and does not directly select a method for the user. It means that having some irrelevant papers analyzed in the whole process will not lead to a completely different result. Moreover, no information is lost, so the trends can be recomputed after filtering if necessary.

## 6 Conclusion

When the experiments are examined, the SARBOLD-LLM produces robust results concerning OpenAI performance for different problem and solution domains in its current state. In terms of the trend, up-to-date usage, and popularity of solution methods, SARBOLD-LLM quickly produces rich and advantageous information for the user. In addition, the recommended keyword selection scheme offers a very flexible structure in choosing the problem domain and solution approach for any use case.

The SARBOLD-LLM completes work in a few hours, which takes a week or more with a manual literature review (selecting AI methods from the title and abstract of 92 papers). It is more suitable for engineers as it proposes methods and trends without adding pros and cons. This limits knowledge accumulation but can be used as a guide for future implementation.

Several prior studies focusing on proposing solution approaches aim to decrease the time and effort spent, emphasizing automation. In alignment with these objectives, SARBOLD-LLM employs a keyword selection strategy to ensure targeted searches within relevant problem and solution domains. Moreover, it incorporates trend, popularity, and relevancy analyzes to derive decision-making insights regarding the optimal solution techniques for various use cases. The research also highlights other outcomes, including sensitivity analyzes conducted for Scopus and OpenAI, and performance results obtained for OpenAI.

### 6.1 Future Work

Due to the nature of the underlying problem, certain processes are technically more difficult to automate than others [5]. In its current form, the SARBOLD-LLM still needs a human to perform the keyword selection, check the results given by the query, classify the found methods, and validate the robustness of the solution. For future work, it would be of high value to remove the need for human intervention while presenting results that signify the trade-off for the different automated decisions. Our study towards automating these tasks is currently underway.

Simultaneously, employing versions from the updated suite of large language models, such as OpenAI's GPT-4,[7] and exploring other databases (like Web of Science, PubMed, IEEE Xplore, etc.) are also future works. Besides, open-source alternatives to GPT-3 or GPT-4, such as GPT-NeoX-20B [75] and GPT-J [76], will be implemented to help in cutting costs.

---

[7] https://openai.com/gpt-4

The sensitivity analysis is split into two parts: queries and prompts. Queries highly depend on the keyword selection scheme and should be studied together. However, reasonably an automatic sensitivity analysis can be made using some variants of the initial query, like using quotation marks instead of brackets or using several forms of the same words. Later, it could be interesting to study the sensitivity concerning synonyms. On the other part, prompts can be analyzed more easily. Indeed, several sentences could be automatically generated concerning the initial one and then tested. The common pool of solutions, or using a scoring-like number of occurrences, could be a robust amicable solution.

Classifying methods is not easy as we want to keep a stratification level from general methods to specific ones. However, as deep learning is already used to classify images, e.g., gaining attention in cancer research [77], a deep learning method could pool different methods together and reduce the number of methods used like YOLO-v2, YOLOv4-tiny, etc. Without any logical pooling, a simple clustering approach based on the text, such as DBSCAN, can be used to make an automatic pooling for a sufficiently big set of methods extracted. However, if we want to automatically match a specific taxonomy, another method will be needed.

Currently, the SARBOLD-LLM only checks the title, abstract, and keywords for the solution approach determination. For certain papers, the specifics of the method are only introduced later in the paper, e.g., for hybrid methods. Consequently, an important extension will be to determine the applied method of a paper from the entirety of a paper.

As we are only providing trends about uses in the literature, the end user still needs to look at surveys to understand the pros and cons of the proposed methods. A future enhancement would be to add a module explaining the pros and cons of each recommended method.

Furthermore, it is planned to convert Python codes to a graphical user interface (GUI) to present automated applications to end users (especially, engineers). Automatic solution approach suggestion simplifies decision-making processes, enhances efficiency, and ensures that decision-makers (end users) have access to relevant, timely, and diverse information to address complex challenges.

Finally, the SARBOLD-LLM can essentially investigate any arbitrary characteristic of the literature rather than only the solution approaches — E.g., identifying problem formulations and varieties therein. Therefore, exploring how to do this manually will greatly benefit the research community.

# Appendix 1 Scopus and OpenAI Results

In Table 6, Scopus results are shown for the initial query stated in Section 4.3. The first column shows the title and DOI information. The second and third columns stand for the electronic identifier and publication year, respectively. C., R., and P. in the last three columns stand for citation number, relevancy value, and popularity value respectively. As mentioned in Section 4.3, articles highlighted in red are manually deleted, and the orange ones that use the AI method are related to the use case but do not specify it in the title and abstract.

In Table 7, OpenAI results for the initial prompt and ground truth methods extracted manually are shown with performance determinants. False-found methods are highlighted in red, and true-found methods are highlighted in green. These performance determinants are utilized to calculate performance metrics stated in Section 4.4.1.

# Appendix 2 OpenAI Performance Results

Below, OpenAI performance results for 55 articles are listed in the same order as Table 7.

- TP = [1, 3, 2, 3, 1, 2, 1, 2, 2, 2, 3, 1, 1, 3, 1, 1, 2, 1, 4, 2, 0, 2, 1, 3, 5, 1, 1, 3, 1, 1, 1, 2, 0, 6, 2, 1, 2, 3, 1, 1, 1,2, 1, 2, 2, 3, 2, 6, 2, 2, 2, 4, 2, 1, 1]
- FP = [0, 1, 0, 0, 0, 2, 0, 2, 1, 0, 0, 1, 2, 2, 0, 1, 3, 1, 0, 0, 3, 0, 1, 2, 3, 1, 0, 0, 0, 0, 1, 2, 0, 0, 1, 2, 0, 0, 1, 1, 0, 0, 1, 0, 1, 1, 1, 2, 0, 1, 1, 1, 1, 5, 2]
- FN = [0, 0, 0, 0, 0, 0, 0, 1, 0, 0, 0, 0, 1, 0, 0, 0, 0, 0, 1, 0, 1, 0, 0, 0, 1, 0, 0, 0, 0, 0, 0, 0, 1, 2, 0, 0, 0, 0, 0, 0,0, 1, 0, 0, 0, 0, 0, 1, 0, 0, 0, 2, 0, 0, 0]
- Precisions = [1, 0.75, 1, 1, 1, 0.5,1, 0.5, 0.6667, 1, 1, 0.5, 0.3334, 0.6, 1, 0.5, 0.4, 0.5, 1, 1, 0, 1, 0.5, 0.6, 0.625, 0.5, 1, 1, 1, 1, 0.5, 0.5, 0, 1, 0.6667, 0.3334, 1, 1, 0.5, 0.5, 1, 1, 0.5, 1, 0.6667, 0.75, 0.6667, 0.75, 1, 0.6667, 0.6667, 0.8, 0.6667, 0.1667, 0.3334] and Average(Precisions) = 0.7111
- Recalls = [1, 1, 1, 1, 1, 1, 1, 0.6667, 1, 1, 1, 1, 0.5, 1, 1, 1, 1, 1, 0.8, 1, 0, 1, 1, 1, 0.8334, 1, 1, 1, 1, 1, 1, 1, 0, 0.75, 1, 1, 1, 1, 1, 1, 1, 0.6667, 1, 1, 1, 1, 1, 0.8571, 1, 1, 1, 0.6667, 1, 1, 1] and Average(Recalls) = 0.9226
- F1-score = [1, 0.8571, 1, 1, 1, 0.6667, 1, 0.5714, 0.8, 1, 1, 0.6667, 0.4, 0.75, 1, 0.6667, 0.5714, 0.6667, 0.8889, 1, 0, 1, 0.6667, 0.75, 0.7143, 0.6667, 1, 1, 1, 1, 0.6667, 0.6667, 0, 0.8571, 0.8, 0.5, 1, 1, 0.6667, 0.6667, 1, 0.8, 0.6667, 1, 0.8, 0.8571, 0.8, 0.8, 1, 0.8, 0.8, 0.7273, 0.8, 0.2857, 0.5] and Average(F1-score) = 0.7775

If all 55 results are considered as a single result pool, then there are 108 TPs, 51 FPs, and 12 FNs. Then precision, recall, and F1-score values are 0.6793, 0.9, and 0.7742, respectively.

When the performance metrics are examined, the OpenAI presents good performance for the manually generated ground truths.

## Appendix 3 OpenAI Sensitivity Results

In Tables 8, 9 and 10, missing and extra/different methods are given regarding the initial prompt. If there is no missing or extra/different method name, it is expressed by "X".

## Appendix 4 Extracted AI Methods and Post-Analyzes

In Table 11, how many times a method is mentioned in the articles is found according to years where the "Papers" column stands for this. The relevancy and popularity sums are written next to the "Papers" column where "Rel." and "Pop." stand for relevancy and popularity, respectively. The total number of articles used is 55 that are not filtered and not general in Table 6. Methods are classified by their occurrence number and their similar ones as described below. Of course, the classification of methods can be done in different ways and at different levels. They are classified to get a more compact overview of the results. The "true general found" results are not included. The methods that are "true found" and mentioned in at least 2 articles are shown.

In the classes listed below, after each method, it is written that it is employed in how many papers total, how many times it is used in which years, and the total relevancy and popularity metrics according to these years.

**Class 1 (Artificial neural networks)**: Paraconsistent Artificial Neural Network (PANN) (×1; 2014, 0, 0.6), Artificial Neural Network (ANN) (×6; 2014, 1, 2.7; 2015, 1, 1; 2016, 0, 6; 2017, 1, 0.7143; 2021, 0, 4.6667; 2023, 0, 2), Probabilistic Neural Network (PNN) (×2; 2015, 0, 0.4444; 2017, 0, 3.2857), Multi-Layer Feed-forward Neural Network (MFFNN) (×1; 2016, 0, 1.125), Neural Networks (×6; 2017×2, 1, 3.8572; 2018, 0, 4; 2019, 0, 0.2; 2020, 1, 0.25; 2023, 0, 0), Perceptron (×1; 2020, 1, 3.75), Back-Propagation Perceptron (×1; 2020, 1, 3.75), Fully Connected Network (FCN) (×1; 2022, 0, 1.5).

Table 6  Journals found regarding Scopus API search

| Title, DOI | EID | Year | C. | R. | P. |
|---|---|---|---|---|---|
| Nevus and melanoma Paraconsistent classification, *10.3233/978-1-61499-474-9-244* | 2-s2.0-84918834255 | 2014 | 6 | 0 | 0.6 |
| A computer-aided diagnostic tool for melanoma, 10.1109/CSCI.2014.26 | 2-s2.0-84902660528 | 2014 | 3 | 1 | 0.3 |
| Cancer therapy prognosis using quantitative ultrasound spectroscopy and a kernel-based metric, *10.1117/12.2043516* | 2-s2.0-84902106481 | 2014 | 8 | 0 | 0.8 |
| Detection of pigment network in dermoscopy images using supervised machine learning and structural analysis, 10.1016/j.compbiomed.2013.11.002 | 2-s2.0-84891153909 | 2014 | 67 | 0 | 6.7 |
| Hybrid genetic algorithm - Artificial neural network classifier for skin cancer detection, *10.1109/ICCICCT.2014.6993162* | 2-s2.0-84891153909 | 2014 | 27 | 1 | 2.7 |
| Implementing DEWA Framework for Early Diagnosis of Melanoma, *10.1016/j.procs.2015.07.555* | 2-s2.0-84948392681 | 2015 | 6 | 0 | 0.6667 |
| Detection of melanoma through image recognition and artificial neural networks, *10.1007/978-3-319-19387-8_204* | 2-s2.0-84944318438 | 2015 | 9 | 1 | 1 |
| Design of a decision support system, trained on GPU, for assisting melanoma | 2-s2.0-84983425726 | 2015 | 4 | 0 | 0.4444 |

Table 6 (continued)

| | | | | | |
|---|---|---|---|---|---|
| diagnosis in dermatoscopy images, *10.1088/1742-6596/633/1/012079* | | | | | |
| Computerized Diagnosis of Melanocytic Lesions Based on the ABCD Method, *10.1109/CLEI.2015.7360029* | 2-s2.0-84961903068 | 2015 | 10 | 0 | 1.1111 |
| Classification of melanoma lesions using sparse coded features and random forests, *10.1117/12.2216973* | 2-s2.0-84988890309 | 2016 | 17 | 1 | 2.125 |
| Slide-specific models for segmentation of differently stained digital histopathology whole slide images, *10.1117/12.2208620* | 2-s2.0-84981719831 | 2016 | 21 | 0 | 2.625 |
| Melanoma image processing and analysis for decision support systems | 2-s2.0-84960402804 | 2016 | 0 | 1 | 0 |
| Validation of a Skin-Lesion Image-Matching Algorithm Based on Computer Vision Technology, *10.1089/tmj.2014.0249* | 2-s2.0-84954436777 | 2016 | 13 | 0 | 1.625 |
| Characterization of melanomas using a variety of features, *10.1109/TIPTEKNO.2015.7374612* | 2-s2.0-84964265510 | 2016 | 0 | 1 | 0 |
| Supervised classification of dermoscopic images using optimized fuzzy clustering based Multi-Layer Feed-forward Neural Network, *10.1109/IC4.2015.7375719* | 2-s2.0-84962792167 | 2016 | 9 | 0 | 1.125 |
| Melanoma detection and classification using SVM based decision support system, *10.1109/INDICON.2015.7443447* | 2-s2.0-84994285804 | 2016 | 14 | 0 | 1.75 |
| Machine Learning Methods for Binary and Multiclass Classification of Melanoma Thickness From Dermoscopic Images, *10.1109/TMI.2015.2506270* | 2-s2.0-84963804529 | 2016 | 48 | 0 | 6 |
| ATLAAS: An automatic decision tree-based learning algorithm for advanced image segmentation in positron emission tomography, *10.1088/0031-9155/61/13/4855* | 2-s2.0-84976426664 | 2016 | 33 | 0 | 4.125 |
| I3DermoscopyApp: Hacking Melanoma thanks to IoT technologies, *10.24251/HICSS.2017.434* | 2-s2.0-85038823793 | 2017 | 4 | 1 | 0.5714 |
| EIMES 3D: An innovative medical images analysis tool to support diagnostic and surgical intervention, *10.1016/j.procs.2017.06.122* | 2-s2.0-85028633382 | 2017 | 23 | 0 | 3.2857 |

**Table 6** (continued)

| Title | ID | Year | | | |
|---|---|---|---|---|---|
| Intelligent system supporting diagnosis of malignant melanoma, *10.1007/978-3-319-60699-6 79* | 2-s2.0-85021253325 | 2017 | 12 | 0 | 1.7143 |
| Dermoscopic feature analysis for melanoma recognition and prevention, *10.1109/INTECH.2016.7845044* | 2-s2.0-85015305877 | 2017 | 8 | 0 | 1.1429 |
| High-level features for automatic skin lesions neural network based classification, *10.1109/IPAS.2016.7880148* | 2-s2.0-85018542177 | 2017 | 19 | 0 | 2.7143 |
| An efficient machine learning approach for the detection of melanoma using dermoscopic images, *10.1109/C-CODE.2017.7918949* | 2-s2.0-85020287797 | 2017 | 64 | 0 | 9.1429 |
| The Rise of Radiomics and Implications for Oncologic Management, *10.1093/jnci/djx055* | 2-s2.0-85021849026 | 2017 | 91 | 0 | 13 |
| Adaptable pattern recognition system for discriminating Melanocytic Nevi from Malignant Melanomas using plain photography images from different image databases, *10.1016/j.ijmedinf.2017.05.016* | 2-s2.0-85019985302 | 2017 | 23 | 0 | 3.2857 |
| Detection of skin cancer 'Melanoma' through computer vision, *10.1109/INTERCON.2017.8079674* | 2-s2.0-85039989899 | 2017 | 8 | 1 | 1.1429 |
| A smart dermoscope design using artificial neural network, *10.1109/IDAP.2017.8090211* | 2-s2.0-85039912655 | 2017 | 5 | 1 | 0.7143 |
| Miaquant, a novel system for automatic segmentation, measurement, and localization comparison of different biomarkers from serialized histological slices, *10.4081/ejh.2017.2838* | 2-s2.0-85036544923 | 2017 | 10 | 0 | 1.4286 |
| Dermoscopic image analysis using pattern recognition techniques from region of interest (ROI) for detection of melanoma, *10.1109/ICMLC.2017.8107759* | 2-s2.0-85042474953 | 2017 | 2 | 1 | 0.2857 |
| Cosmetic oncology: Innocent mole or malignant melanoma? Subjective assessments, objective semiology and aided diagnosis | 2-s2.0-85103213412 | 2018 | 0 | 1 | 0 |
| Simulation and Synthesis in Medical Imaging, *10.1109/TMI.2018.2800298* | 2-s2.0-85042927927 | 2018 | 60 | 0 | 10 |

Table 6 (continued)

| | | | | | |
|---|---|---|---|---|---|
| Feature selection using sequential backward method in melanoma recognition, *10.1109/ICECCO.2017.8333341* | 2-s2.0-85050502789 | 2018 | 12 | 1 | 2 |
| Machine learning–based diagnosis of melanoma using macro images, *10.1002/cnm.2953* | 2-s2.0-85042181700 | 2018 | 24 | 0 | 4 |
| Computer aided early detection and classification of malignant melanoma, *10.1109/CICN.2018.8864942* | 2-s2.0-85074209854 | 2018 | 0 | 1 | 0 |
| Dermoscopic assisted diagnosis in melanoma: Reviewing results, optimizing methodologies and quantifying empirical guidelines, *10.1016/j.knosys.2018.05.016* | 2-s2.0-85048760650 | 2018 | 25 | 1 | 4.1667 |
| Estimation of Illumination Map from Dermoscopy Images for Extracting Differential Structures Using Gabor Local Mesh Patterns, *10.1109/CSCI.2017.72* | 2-s2.0-85060645376 | 2018 | 0 | 0 | 0 |
| A computer aided diagnosis system for skin cancer detection, *10.1007/978-3-030-05532-5_42* | 2-s2.0-85059759729 | 2019 | 6 | 1 | 1.2 |
| The Continuing Evolution of Molecular Functional Imaging in Clinical Oncology: The Road to Precision Medicine and Radiogenomics (Part I), *10.1007/s40291-018-0366-4* | 2-s2.0-85056306068 | 2019 | 15 | 1 | 3 |
| The Continuing Evolution of Molecular Functional Imaging in Clinical Oncology: The Road to Precision Medicine and Radiogenomics (Part II), *10.1007/s40291-018-0367-3* | 2-s2.0-85055963110 | 2019 | 6 | 1 | 1.2 |
| Towards a modular decision support system for radiomics: A case study on rectal cancer, *10.1016/j.artmed.2018.09.003* | 2-s2.0-85054192769 | 2019 | 29 | 0 | 5.8 |
| Radiomics to predict prostate canceraggressiveness: A preliminary study, *10.1109/BIBE.2019.00181* | 2-s2.0-85077975055 | 2019 | 4 | 1 | 0.8 |
| Basis and perspectives of artificial intelligence in radiation therapy, *10.1016/j.canrad.2019.08.005* | 2-s2.0-85073726145 | 2019 | 1 | 0 | 0.2 |
| Melanoma detection using an objective system based on multiple connected neural networks, *10.1109/ACCESS.2020.3028248* | 2-s2.0-85099879408 | 2020 | 15 | 1 | 3.75 |

**Table 6** (continued)

| | | | | | |
|---|---|---|---|---|---|
| Skin cancer classification computer system development with deep learning | 2-s2.0-85091193990 | 2020 | 0 | 1 | 0 |
| Methods of artificial intelligence and their application in imaging diagnostics | 2-s2.0-85086354606 | 2020 | 2 | 1 | 0.5 |
| Automatic classification of melanocytic skin tumors based on hyperparameters optimized by cross-validation using support vector machines, *10.1117/12.2542161* | 2-s2.0-85081629940 | 2020 | 0 | 1 | 0 |
| Clinical implementation of AI technologies will require interpretable AI models, 10.1002/mp.13891 | 2-s2.0-85075297765 | 2020 | 50 | 0 | 12.5 |
| Utilizing computer vision, clustering and neural networks for melanoma categorization, *10.1145/3374135.3385327* | 2-s2.0-85086182046 | 2020 | 1 | 1 | 0.25 |
| Artificial intelligence in oncology, *10.1111/cas.14377* | 2-s2.0-85082062276 | 2020 | 82 | 0 | 20.5 |
| Artificial intelligence radiogenomics for advancing precision and effectiveness in oncologic care (Review), *10.3892/ijo.2020.5063* | 2-s2.0-85084500870 | 2020 | 29 | 0 | 7.25 |
| A preliminary study on machine learning-based evaluation of static and dynamic fet-pet for the detection of pseudoprogression in patients with idh-wildtype glioblastoma, *10.3390/cancers12113080* | 2-s2.0-85093943094 | 2020 | 18 | 0 | 4.5 |
| U-Net-based medical image segmentation algorithm, *10.1109/WCSP52459.2021.9613447* | 2-s2.0-85123369429 | 2021 | 0 | 1 | 0 |
| Skin Pathology Detection Using Artificial Intelligence, *10.1109/ISPCC53510.2021.9609516* | 2-s2.0-85123014841 | 2021 | 5 | 1 | 1.6667 |
| Developing a Recognition System for Diagnosing Melanoma Skin Lesions Using Artificial Intelligence Algorithms, *10.1155/2021/9998379* | 2-s2.0-85107175102 | 2021 | 14 | 0 | 4.6667 |
| Microfluidic Co-Culture Models for Dissecting the Immune Response in in vitro Tumor Microenvironments, *10.3791/61895* | 2-s2.0-85138248078 | 2021 | 8 | 1 | 2.6667 |
| Quantitative Whole Slide Assessment of Tumor-Infiltrating CD8-Positive Lymphocytes in ER-Positive Breast | 2-s2.0-85096782074 | 2021 | 11 | 0 | 3.6667 |

**Table 6** (continued)

| | | | | | |
|---|---|---|---|---|---|
| Cancer in Relation to Clinical Outcome, *10.1109/JBHI.2020.3003475* | | | | | |
| Artificial intelligence: Deep learning in oncological radiomics and challenges of interpretability and data harmonization, *10.1016/j.ejmp.2021.03.009* | 2-s2.0-85102886091 | 2021 | 55 | 1 | 18.3333 |
| Quantitative PET in the 2020s: A roadmap, *10.1088/1361-6560/abd4f7* | 2-s2.0-85103515923 | 2021 | 23 | 0 | 7.6667 |
| Virtual reality and artificial intelligence for 3-dimensional planning of lung segmentectomies, *10.1016/j.xjtc.2021.03.016* | 2-s2.0-85103937126 | 2021 | 22 | 0 | 7.3333 |
| Artificial Intelligence-based methods in head and neck cancer diagnosis: an overview, *10.1038/s41416-021-01386-x* | 2-s2.0-85104847620 | 2021 | 26 | 0 | 8.6667 |
| Global evolution of research on pulmonary nodules: A bibliometric analysis, *10.2217/fon-2020-0987* | 2-s2.0-85108124165 | 2021 | 2 | 0 | 0.6667 |
| The constantly evolving role of medical image processing in oncology: From traditional medical image processing to imaging biomarkers and radiomics, *10.3390/jimaging7080124* | 2-s2.0-85111942180 | 2021 | 2 | 1 | 0.6667 |
| Automatic contour segmentation of cervical cancer using artificial intelligence, *10.1093/jrr/rrab070* | 2-s2.0-85116348283 | 2021 | 7 | 0 | 2.3333 |
| The Digital Twin: Modular Model-Based Approach to Personalized Medicine, *10.1515/cdbme-2021-2057* | 2-s2.0-85121865792 | 2021 | 2 | 0 | 0.6667 |
| Malignant Melenoma and Atypical Nevus Classification Using Machine Learning with Shape and Assymetric Features, *10.1109/GCAT52182.2021.9587838* | 2-s2.0-85119477680 | 2021 | 1 | 1 | 0.3333 |
| Radiomics in oncology: A practical guide, *10.1148/rg.2021210037* | 2-s2.0-85117052739 | 2021 | 67 | 0 | 22.3333 |
| Artificial intelligence in radiation oncology: A review of its current status and potential application for the radiotherapy workforce, *10.1016/j.radi.2021.07.012* | 2-s2.0-85114341702 | 2021 | 10 | 0 | 3.3333 |
| Artificial Intelligence in Cancer Care: Legal and Regulatory Dimensions, *10.1002/onco.13862* | 2-s2.0-85111119804 | 2021 | 2 | 0 | 0.6667 |
| Model for Estimating the Heterogeneity of the Distribution of Globule Characteristics in Images of Skin | 2-s2.0-85124348701 | 2021 | 1 | 0 | 0.3333 |

**Table 6** (continued)

| | | | | | |
|---|---|---|---|---|---|
| Neoplasms, *10.1007/s11018-022-02003-w* | | | | | |
| Abdominal Computed Tomography Enhanced Image Features under an Automatic Segmentation Algorithm in Identification of Gastric Cancer and Gastric Lymphoma, *10.1155/2022/2259373* | 2-s2.0-85135422352 | 2022 | 0 | 1 | 0 |
| MRI radiomics-based machine learning classification of atypical cartilaginous tumour and grade II chondrosarcoma of long bones, *10.1016/j.ebiom.2021.103757* | 2-s2.0-85121330341 | 2022 | 17 | 0 | 8.5 |
| Artificial intelligence-based classification of bone tumors in the proximal femur on plain radiographs: System development and validation, *10.1371/journal.pone.0264140* | 2-s2.0-85125337919 | 2022 | 4 | 0 | 2 |
| Segmentation of skin lesions image based on U-Net + +, *10.1007/s11042-022-12067-z* | 2-s2.0-85124261031 | 2022 | 3 | 0 | 1.5 |
| Brain Tumor Imaging: Applications of Artificial Intelligence, *10.1053/j.sult.2022.02.005* | 2-s2.0-85125529381 | 2022 | 4 | 1 | 2 |
| Profiling the most highly cited scholars from China: Who they are. To what extent they are interdisciplinary, *10.3145/epi.2022.jul.08* | 2-s2.0-85137378624 | 2022 | 0 | 1 | 0 |
| Clinical Validation of a Deep-Learning Segmentation Software in Head and Neck: An Early Analysis in a Developing Radiation Oncology Center, *10.3390/ijerph19159057* | 2-s2.0-85135382796 | 2022 | 4 | 0 | 2 |
| Radiomics: a primer on high-throughput image phenotyping, *10.1007/s00261-021-03254-x* | 2-s2.0-85113512614 | 2022 | 20 | 0 | 10 |
| Clinical application of deep learning-based synthetic CT from real MRI to improve dose planning accuracy in Gamma Knife radiosurgery: a proof of concept study, *10.1007/s13534-022-00227-x* | 2-s2.0-85131873486 | 2022 | 0 | 0 | 0 |
| BLSNet: Skin lesion detection and classification using broad learning system with incremental learning algorithm, *10.1111/exsy.12938* | 2-s2.0-85124549532 | 2022 | 3 | 1 | 1.5 |

**Table 6** (continued)

| | | | | | |
|---|---|---|---|---|---|
| Fully Automated, Semantic Segmentation of Whole-Body <sup>18</sup>F-FDG PET/CT Images Based on Data-Centric Artificial Intelligence, *10.2967/jnumed.122.264063* | 2-s2.0-85139515590 | 2022 | 3 | 0 | 1.5 |
| A deep image-to-image network organ segmentation algorithm for radiation treatment planning: principles and evaluation, *10.1186/s13014-022-02102-6* | 2-s2.0-85134588716 | 2022 | 2 | 0 | 1 |
| Detection of Melanomas Using Ensembles of Deep Convolutional Neural Networks, *10.1109/ATEE58038.2023.10108394* | 2-s2.0-85159074452 | 2023 | 0 | 0 | 0 |
| Transfer learning with different modified convolutional neural network models for classifying digital mammograms utilizing Local Dataset | 2-s2.0-85148402129 | 2023 | 0 | 1 | 0 |
| Neural Network in the Analysis of the MR Signal as an Image Segmentation Tool for the Determination of T1 and T2 Relaxation Times with Application to Cancer Cell Culture, *10.3390/ijms24021554* | 2-s2.0-85146500215 | 2023 | 0 | 0 | 0 |
| Intraclass Clustering-Based CNN Approach for Detection of Malignant Melanoma, *10.3390/s23020926* | 2-s2.0-85146428707 | 2023 | 0 | 0 | 0 |
| Unsupervised Learning Composite Network to Reduce Training Cost of Deep Learning Model for Colorectal Cancer Diagnosis, *10.1109/JTEHM.2022.3224021* | 2-s2.0-85144032371 | 2023 | 0 | 0 | 0 |
| Detecting skin lesions fusing handcrafted features in image network ensembles, *10.1007/s11042-022-13046-0* | 2-s2.0-85131537962 | 2023 | 2 | 0 | 2 |
| An IoMT-Based Melanoma Lesion Segmentation Using Conditional Generative Adversarial Networks, *10.3390/s23073548* | 2-s2.0-85152350497 | 2023 | 0 | 0 | 0 |
| Multi-strategy ant colony optimization for multi-level image segmentation: Case study of melanoma, *10.1016/j.bspc.2023.104647* | 2-s2.0-85147848611 | 2023 | 1 | 1 | 1 |
| Application of 3D-reconstruction and artificial intelligence for complete mesocolic excision and D3 lymphadenectomy in colon cancer, *10.1016/j.ciresp.2022.10.023* | 2-s2.0-85144536545 | 2023 | 0 | 1 | 0 |
| Detection of melanoma with hybrid learning method by removing hair from dermoscopic images using image processing techniques and wavelet transform, *10.1016/j.bspc.2023.104729* | 2-s2.0-85148874931 | 2023 | 0 | 1 | 0 |

Table 7 OpenAI and manually found results for employed AI methods in each article

| EID | Methods (OpenAI) | Methods (manual) | Performance Determinants |
|---|---|---|---|
| 2-s2.0-84918834255 | Paraconsistent Artificial Neural Network (PANN) | Paraconsistent Artificial Neural Network (PANN) | Total manual: 1<br>True found: 1<br>False found: 0<br>Missing: 0<br>True general: 0 |
| 2-s2.0-84902106481 | Kernel-based metric, Hilbert-Schmidt independence criterion (HSIC), reproducing kernel Hilbert space (RKHS), k-nearest-neighbor (k-NN) classifier | Hilbert-Schmidt independence criterion (HSIC), reproducing kernel Hilbert space (RKHS), k-nearest neighbor (k-NN) classifier | Total manual: 3<br>True found: 3<br>False found: 1<br>Missing: 0<br>True general: 0 |
| 2-s2.0-84921642433 | Artificial Neural Network, Genetic Algorithm | Artificial neural network, genetic algorithm | Total manual: 2<br>True found: 2<br>False found: 0<br>Missing: 0<br>True general: 0 |
| 2-s2.0-84948392681 | Connected Component Labelling, K-Means, Morphological Filter | Connected component labelling, K-means, morphological filter | Total manual: 3<br>True found: 3<br>False found: 0<br>Missing: 0<br>True general: 0 |
| 2-s2.0-84944318438 | Artificial Neural Networks | Artificial neural network | Total manual: 1<br>True found: 1<br>False found: 0<br>Missing: 0<br>True general: 0 |

**Table 7** (continued)

| | | | |
|---|---|---|---|
| 2-s2.0-84983425726 | Probabilistic Neural Network Classifier, Dull Razor algorithm, Level Sets, Automated Thresholding Approach | Dull Razor algorithm, Probabilistic Neural Network Classifier | Total manual: 2<br>True found: 2<br>False found: 2<br>Missing: 0<br>True general: 0 |
| 2-s2.0-84961903068 | Support Vector Machine | Support Vector Machine | Total manual: 1<br>True found: 1<br>False found: 0<br>Missing: 0<br>True general: 0 |
| 2-s2.0-84988890309 | Machine learning, Image processing, Random Forests, Sparse Coding | Random Forests, Sparse Coding, SIFT | Total manual: 3<br>True found: 2<br>False found: 0<br>Missing: 1<br>True general: 2 |
| 2-s2.0-84962792167 | Supervised classification, Multi-Layer Feed-forward Neural Network, Genetically Optimized Fuzzy C-means clustering | Multi-Layer Feed-forward Neural Network, Genetically Optimized Fuzzy C-means clustering | Total manual: 2<br>True found: 2<br>False found: 0<br>Missing: 0<br>True general: 1 |
| 2-s2.0-84994285804 | Support Vector Machine (SVM), Sequential Minimal Optimization (SMO) | Support Vector Machine (SVM), Sequential Minimal Optimization (SMO) | Total manual: 2<br>True found: 2<br>False found: 0<br>Missing: 0<br>True general: 0 |
| 2-s2.0-84963804529 | Artificial Neural Networks, Logistic Regression, LIPU | Artificial neural networks, Logistic regression, Logistic regression using Initial variables and Product Units (LIPU) | Total manual: 3<br>True found: 3<br>False found: 0<br>Missing: 0<br>True general: 0 |
| 2-s2.0-84976426664 | Supervised Machine Learning, Decision Trees | Decision tree | Total manual: 1<br>True found: 1<br>False found: 0<br>Missing: 0<br>True general: 1 |

**Table 7** (continued)

| | | | |
|---|---|---|---|
| 2-s2.0-85018542177 | Neural Network based classification, Shape Characterization, Color and Texture Features | Neural network classifier, semantic analysis | Total manual: 2<br>True found: 1<br>False found: 2<br>Missing: 1<br>True general: 0 |
| 2-s2.0-85019985302 | Probabilistic Neural Network (PNN), Exhaustive Search Features Selection, Leave-one-out (LOO), External Cross-validation (ECV) | Probabilistic Neural Network (PNN) classifier, leave-one-out (LOO), external cross-validation (ECV) | Total manual: 3<br>True found: 3<br>False found: 2<br>Missing: 0<br>True general: 0 |
| 2-s2.0-85039989899 | Neural Networks | Neural networks | Total manual: 1<br>True found: 1<br>False found: 0<br>Missing: 0<br>True general: 0 |
| 2-s2.0-85039912655 | Artificial Neural Network, Learning Algorithm | Artificial Neural Network | Total manual: 1<br>True found: 1<br>False found: 0<br>Missing: 0<br>True general: 1 |
| 2-s2.0-85042474953 | Pattern recognition, Adaptive thresholding, Morphological operators, Texture features, Color features | Pattern recognition, Adaptive thresholding | Total manual: 2<br>True found: 2<br>False found: 3<br>Missing: 0<br>True general: 0 |
| 2-s2.0-85050502789 | Machine Learning, k-Nearest Neighbors | k-Nearest Neighbors algorithm | Total manual: 1<br>True found: 1<br>False found: 0<br>Missing: 0<br>True general: 1 |
| 2-s2.0-85042181700 | Support Vector Machine, Random Forest, Neural Network, Fast Discriminative Mixed-Membership–Based Naive Bayesian Classifiers | Support vector machine, random forest, neural network, fast discriminative mixed-membership–based naive Bayesian classifiers, information theory | Total manual: 5<br>True found: 4<br>False found: 0<br>Missing: 1<br>True general: 0 |

**Table 7** (continued)

| | | | |
|---|---|---|---|
| 2-s2.0-85074209854 | GLCM, SVM | SVM, GLCM, Note: Grey Level Co-Occurrence Matrix | Total manual: 2<br>True found: 2<br>False found: 0<br>Missing: 0<br>True general: 0 |
| 2-s2.0-85048760650 | Machine Learning, Digital Image Processing, Feature Selection | Decision tree | Total manual: 1<br>True found: 0<br>False found: 1<br>Missing: 1<br>True general: 2 |
| 2-s2.0-85060645376 | Gabor filtering, Local Mesh Patterns | Gabor filtering, Local mesh patterns | Total manual: 2<br>True found: 2<br>False found: 0<br>Missing: 0<br>True general: 0 |
| 2-s2.0-85059759729 | Machine learning algorithms, Support Vector Machines | Support vector machines | Total manual: 1<br>True found: 1<br>False found: 0<br>Missing: 0<br>True general: 1 |
| 2-s2.0-85073726145 | Connectionism, Logics, Neural Networks, General Adversial Networks, Deep Learning | Neural network, General adversial networks, deep learning | Total manual: 3<br>True found: 3<br>False found: 2<br>Missing: 0<br>True general: 0 |
| 2-s2.0-85099879408 | Perceptron, Color Local Binary Patterns, Color Histograms of Oriented Gradients, Generative Adversarial Network, ABCD Rule, ResNet, AlexNet, Back-Propagation Perceptron | Neural network, perceptron, generative adversarial network, ResNet, AlexNet, back-propagation perceptron | Total manual: 6<br>True found: 5<br>False found: 3<br>Missing: 1<br>True general: 0 |
| 2-s2.0-85091193990 | Artificial Intelligence, Deep Learning | Deep learning | Total manual: 1<br>True found: 1<br>False found: 0<br>Missing: 0<br>True general: 1 |

**Table 7** (continued)

| | | | |
|---|---|---|---|
| 2-s2.0-85081629940 | Support Vector Machines | Support vector machines | Total manual: 1<br>True found: 1<br>False found: 0<br>Missing: 0<br>True general: 0 |
| 2-s2.0-85086182046 | Computer Vision, Clustering, Neural Networks | Computer Vision, Clustering, Neural Networks | Total manual: 3<br>True found: 3<br>False found: 0<br>Missing: 0<br>True general: 0 |
| 2-s2.0-85082062276 | Deep learning | Deep learning | Total manual: 1<br>True found: 1<br>False found: 0<br>Missing: 0<br>True general: 0 |
| 2-s2.0-85084500870 | Deep learning | Deep learning | Total manual: 1<br>True found: 1<br>False found: 0<br>Missing: 0<br>True general: 0 |
| 2-s2.0-85093943094 | Machine Learning, Linear Discriminant Analysis | Linear Discriminant Analysis | Total manual: 1<br>True found: 1<br>False found: 0<br>Missing: 0<br>True general: 1 |
| 2-s2.0-85123369429 | U-Net, Deep Learning, Image Segmentation, Artificial Intelligence | U-Net, Deep learning | Total manual: 2<br>True found: 2<br>False found: 0<br>Missing: 0<br>True general: 2 |
| 2-s2.0-85123014841 | | CNN | Total manual: 1<br>True found: 0<br>False found: 0<br>Missing: 1<br>True general: 0 |

**Table 7** (continued)

| | | | |
|---|---|---|---|
| 2-s2.0-85107175102 | Artificial Neural Network (ANNs), Local Binary Pattern (LBP), Gray Level Co-occurrence Matrix (GLCM), Convolutional Neural Network (CNNs), AlexNet, ResNet50 | Deep learning, active contour method, Local Binary Pattern (LBP), Gray Level Co-occurrence Matrix (GLCM), artificial neural network (ANNs), convolutional neural network (CNNs), AlexNet, ResNet50 | Total manual: 8<br>True found: 6<br>False found: 0<br>Missing: 2<br>True general: 0 |
| 2-s2.0-85096782074 | Deep learning, Image registration, Deep learning-based nucleus detection | Deep learning, Image registration | Total manual: 2<br>True found: 2<br>False found: 1<br>Missing: 0<br>True general: 0 |
| 2-s2.0-85108124165 | Deep Learning, Artificial Intelligence, Machine Learning | Deep learning | Total manual: 1<br>True found: 1<br>False found: 0<br>Missing: 0<br>True general: 2 |
| 2-s2.0-85116348283 | 2D U-Net, 3D U-Net | 2D U-Net, 3D U-Net | Total manual: 2<br>True found: 2<br>False found: 0<br>Missing: 0<br>True general: 0 |
| 2-s2.0-85119477680 | SVM, KNN, Ensemble Learning | SVM, KNN, ensemble learning | Total manual: 3<br>True found: 3<br>False found: 0<br>Missing: 0<br>True general: 0 |
| 2-s2.0-85135422352 | OTSU threshold segmentation, artificial intelligence algorithms | OTSU | Total manual: 1<br>True found: 1<br>False found: 0<br>Missing: 0<br>True general: 1 |
| 2-s2.0-85121330341 | Extra Trees Classifier, Machine Learning | Extra Trees Classifier | Total manual: 1<br>True found: 1<br>False found: 0<br>Missing: 0<br>True general: 1 |

**Table 7** (continued)

| | | | |
|---|---|---|---|
| 2-s2.0-85125337919 | Convolutional Neural Network (CNN) algorithms | Convolutional neural network (CNN) | Total manual: 1<br>True found: 1<br>False found: 0<br>Missing: 0<br>True general: 0 |
| 2-s2.0-85124261031 | Fully Connected Networks (FCNs) and U-Net | U-Net, fully connected networks (FCNs), U-Net++ | Total manual: 3<br>True found: 2<br>False found: 0<br>Missing: 1<br>True general: 0 |
| 2-s2.0-85135382796 | Deep-learning, auto-segmentation | Deep learning | Total manual: 1<br>True found: 1<br>False found: 1<br>Missing: 0<br>True general: 0 |
| 2-s2.0-85131873486 | Deep learning, Convolution algorithm | Deep learning, Convolution algorithm | Total manual: 2<br>True found: 2<br>False found: 0<br>Missing: 0<br>True general: 0 |
| 2-s2.0-85124549532 | Deep Learning (DL), Broad Learning System (BLS), Incremental Learning Algorithm | Deep learning, incremental learning algorithm | Total manual: 2<br>True found: 2<br>False found: 1<br>Missing: 0<br>True general: 0 |
| 2-s2.0-85134588716 | Deep Reinforcement Learning, Deep Image-to-Image Network (DI2IN), Convolutional Encoder-Decoder Architecture, Multi-Level Feature Concatenation | Deep reinforcement learning, convolutional encoder-decoder architecture, multi-level feature concatenation | Total manual: 3<br>True found: 3<br>False found: 1<br>Missing: 0<br>True general: 0 |
| 2-s2.0-85159074452 | Deep Convolutional Neural Networks, Fusion of the decisions of several neural networks, Horizontal Voting | Deep convolutional neural networks, horizontal voting ensemble | Total manual: 2<br>True found: 2<br>False found: 1<br>Missing: 0<br>True general: 0 |

**Table 7** (continued)

| | | | |
|---|---|---|---|
| 2-s2.0-85148402129 | Transfer learning, Convolutional Neural Network, Machine Learning Algorithms, Contrast Limited Adaptive Histogram Equalization (CLAHE), Data Augmentation, NASNetLarge, DenseNet169, InceptionResNetV2 | Transfer learning, convolutional neural network, NASNetLarge, DenseNet169, InceptionResNetV2, data augmentation, fine tuning | Total manual: 7<br>True found: 6<br>False found: 1<br>Missing: 1<br>True general: 1 |
| 2-s2.0-85146500215 | Neural Networks, Deep Learning | Neural network, deep learning | Total manual: 2<br>True found: 2<br>False found: 0<br>Missing: 0<br>True general: 0 |
| 2-s2.0-85146428707 | Artificial Intelligence (AI), Convolutional Neural Network (CNN), Intraclass Clustering | Convolutional neural network (CNN), Intraclass Clustering | Total manual: 2<br>True found: 2<br>False found: 0<br>Missing: 0<br>True general: 1 |
| 2-s2.0-85144032371 | Unsupervised Learning, K-means Clustering Algorithm, Deep Learning | Deep learning, K-means clustering | Total manual: 2<br>True found: 2<br>False found: 0<br>Missing: 0<br>True general: 1 |
| 2-s2.0-85131537962 | Artificial Intelligence, Deep Learning, EfficientNets, Artificial Neural Network, Majority Soft Voting | Deep learning, EfficientNets, artificial neural network, ensemble learning, majority soft voting, transfer learning | Total manual: 6<br>True found: 4<br>False found: 0<br>Missing: 2<br>True general: 1 |
| 2-s2.0-85152350497 | Artificial Intelligence, Deep Learning, Conditional Generative Adversarial Network (cGAN) | Conditional Generative Adversarial Network (cGAN), generative deep learning | Total manual: 2<br>True found: 2<br>False found: 0<br>Missing: 0<br>True general: 1 |
| 2-s2.0-85147848611 | Ant Colony Optimization, Sine Cosine Strategy, Disperse Foraging Strategy, Specular Reflection Learning Strategy, Non-Local Mean Strategy, 2D Kapur's Entropy Strategy | Ant colony optimization | Total manual: 1<br>True found: 1<br>False found: 5<br>Missing: 0<br>True general: 0 |
| 2-s2.0-85148874931 | Artificial Intelligence, Deep Learning, Machine Learning | Deep learning | Total manual: 1<br>True found: 1<br>False found: 0<br>Missing: 0<br>True general: 2 |

**Table 8** Missing and extra/different methods for prompts 1 and 2 regarding the initial prompt results

| EID | Prompt 1 | | Prompt 2 | |
|---|---|---|---|---|
| | Missing | Extra or different | Missing | Extra or different |
| 2-s2.0-84918834255 | X | X | X | X |
| 2-s2.0-84902106481 | X | X | X | quantitative ultrasound (QUS), radiofrequency (RF) signals, Euclidean distance |
| 2-s2.0-84921642433 | X | Digital Image processing, GLCM, RGB color feature | X | X |
| 2-s2.0-84,948,392,681 | X | Segmentation, Filtering | X | Segmentation, Filtering, Learning and Non-Learning Methods |
| 2-s2.0-84944318438 | X | Image processing | X | Image processing |
| 2-s2.0-84983425726 | X | Exhaustive search, Leave one out method, GPU card (GeForce 580GTX), CUDA programming framework, C++ programming language | X | Exhaustive search, Leave one out method, CUDA programming framework, C++ programming language |
| 2-s2.0-84961903068 | X | Preprocessing, Segmentation, Feature Extraction, Classification | X | Preprocessing, Segmentation, Feature Extraction, Classification |
| 2-s2.0-84988890309 | Sparse Coding | Pre-processing, Segmentation, Classification, SIFT, Hue, Opponent Angle Histograms, RGB Intensities, Dictionary Learning | Sparse Coding | Pre-processing, Segmentation, Classification, SIFT, Hue, Opponent Angle Histograms, RGB Intensities, Dictionary Learning |
| 2-s2.0-84962792167 | X | Optimized Fuzzy Clustering, Machine Learning | X | Optimized Fuzzy Clustering |
| 2-s2.0-84994285804 | X | Iterative Dilation Method, Feature Vector, SVM Classifier | X | Iterative Dilation Method, SVM Classifier |
| 2-s2.0-84963804529 | X | X | LIPU | Machine Learning, Ordinal Classification |
| 2-s2.0-84976426664 | X | Dice Similarity Coefficient (DSC) | X | Dice Similarity Coefficient (DSC) |
| 2-s2.0-85018542177 | X | Feature Extraction | X | Feature Extraction, ABCD Rule, 7-Point Checklist, Menzies Method, CASH Algorithm |
| 2-s2.0-85019985302 | X | X | X | X |
| 2-s2.0-85039989899 | X | Image Processing, Computer Vision | X | Image Processing |
| 2-s2.0-85039912655 | X | Image Processing Software, Artificial Neural Network Learning Algorithm | X | Image Processing Software, Artificial Neural Network Learning Algorithm |
| 2-s2.0-85042474953 | X | X | X | X |
| 2-s2.0-85050502789 | X | Feature selection, Sequential backward selection, Image processing, Segmentation | X | Image Processing, Feature Extraction, Segmentation, Sequential Backward Selection |

**Table 8** (continued)

| EID | Prompt 1 | | Prompt 2 | |
|---|---|---|---|---|
| | Missing | Extra or different | Missing | Extra or different |
| 2-s2.0-85042181700 | X | Multistage Illumination Compensation, Multimode Segmentation, Information Theory | X | Multistage Illumination Compensation, Multimode Segmentation |
| 2-s2.0-85074209854 | X | Dull-Razor, Image Processing, Automatic Segmentation, Basic Statistical Method | X | Dull-Razor, Image Processing, Automatic Segmentation, Basic Statistical Method |
| 2-s2.0-85048760650 | X | Decision Tree | X | Decision Tree |
| 2-s2.0-85060645376 | X | Statistical estimation | X | Statistical estimation |
| 2-s2.0-85059759729 | X | Image processing techniques | X | Image Processing Techniques |
| 2-s2.0-85073726145 | X | Natural Language Processing, Radiomics | X | Deduction, Induction, Abduction, Radiomics, Natural Language Processing |
| 2-s2.0-85081629940 | X | Cross-Validation, Power Spectral Densities, Gray-Level Co-Occurrence Matrices, Holdout Validation, Stratified Cross-Validation | X | Cross-Validation, Image Processing, Artificial Intelligence |
| 2-s2.0-85091193990 | X | Neural Network Architecture | Artificial Intelligence | X |
| 2-s2.0-85099879408 | X | X | X | X |
| 2-s2.0-85086182046 | X | Artificial Intelligence, Image Processing | X | X |
| 2-s2.0-85082062276 | X | AI, feature extraction | X | X |
| 2-s2.0-85084500870 | X | Radiogenomics, Precision Medicine, Computational Medical Imaging, Molecular Expression | X | X |
| 2-s2.0-85093943094 | X | Receiver-Operating -Characteristic (ROC), Tumor-to-Brain Ratios (TBRmean, TBRmax) | X | Receiver-Operating -Characteristic (ROC) curve |
| 2-s2.0-85123369429 | Image, Segmentation | Path Direct Connection, Dropout Direct Connection, Conv Direct Connection, Constant Scale | Image, Segmentation | Path Direct Connection, Dropout Direct Connection, Conv Direct Connection, Constant Scale |
| 2-s2.0-85107175102 | X | Deep Learning, Transfer Learning | X | Deep Learning |
| 2-s2.0-85123014841 | X | CNN | X | CNN |
| 2-s2.0-85096782074 | X | tissue-type classification algorithm, nucleus detection, immunohistochemistry (IHC) | X | tissue-type classification algorithm, nucleus detection, immunohistochemistry (IHC) |
| 2-s2.0-85108124165 | X | X | X | X |

**Table 8** (continued)

| EID | Prompt 1 | | Prompt 2 | |
|---|---|---|---|---|
| | Missing | Extra or different | Missing | Extra or different |
| 2-s2.0-85116348283 | X | Dice similarity coefficient (DSC), Hausdorff distance (HD) | X | Dice similarity coefficient (DSC), Hausdorff distance (HD) |
| 2-s2.0-85119477680 | X | X | X | Machine Learning |
| 2-s2.0-85135422352 | X | CT Image Processing | X | X |
| 2-s2.0-85121330341 | X | Bidimensional Segmentation, Radiomic Features, Dimensionality Reduction, Class Balancing, 10-Fold Cross-Validation, McNemar's Test | X | Bidimensional Segmentation, Dimensionality Reduction, Class Balancing, 10-Fold Cross-Validation, McNemar's Test |
| 2-s2.0-85125337919 | X | Deep Learning, Receiver Operating Characteristic (ROC), Fivefold Cross-Validation | X | Receiver Operating Characteristic (ROC), Area Under the Receiver Operating Characteristic (AUROC) |
| 2-s2.0-85124261031 | X | U-Net++, Jaccard index | X | U-Net++, Loss Function |
| 2-s2.0-85135382796 | X | Dice similarity coefficient (DSC), Hausdorff distance transform (DT) | X | Dice similarity coefficient (DSC), 95% Hausdorff distance transform (DT) |
| 2-s2.0-85131873486 | X | TMR algorithm, framebased contrast-enhanced T1-weighted MR images, synthetic CT (sCT), mean absolute error (MAE) | X | TMR algorithm, Synthetic CT (sCT), Convolution with rCT (Conv-rCT) plan, Convolution with sCT (Conv-sCT) plan, Mean Absolute Error (MAE) |
| 2-s2.0-85124549532 | X | X | X | X |
| 2-s2.0-85134588716 | Deep Image-to-Image Network (DI2IN) | Dice Similarity Coefficient (DSC), Hausdorff Distance (HD95) | Deep Image-to-Image Network (DI2IN) | Dice Similarity Coefficient (DSC), Hausdorff Distance (HD95) |
| 2-s2.0-85144032371 | X | Artificial Intelligence | X | X |
| 2-s2.0-85,146,428,707 | X | X | X | X |
| 2-s2.0-85,146,500,215 | X | Patternnet, DICOM, MATLAB | X | Patternnet, MATLAB |
| 2-s2.0-85148402129 | Machine Learning Algorithms | Keras, Python | Machine Learning Algorithms | X |
| 2-s2.0-85159074452 | X | X | X | X |
| 2-s2.0-85131537962 | Artificial Intelligence, Deep Learning, EfficientNets, Artificial Neural Network, Majority Soft Voting | X | Artificial Intelligence, Deep Learning | Transfer Learning, Image Data, Handcrafted Lesion Features, Metadata |
| 2-s2.0-85152350497 | X | Computer Vision | X | X |

**Table 8** (continued)

| EID | Prompt 1 | | Prompt 2 | |
|---|---|---|---|---|
| | Missing | Extra or different | Missing | Extra or different |
| 2-s2.0-85147848611 | X | X | X | X |
| 2-s2.0-85148874931 | Artificial Intelligence | Image processing techniques, Wavelet transform | Artificial Intelligence | Image processing techniques, Wavelet transform |

**Table 9** Missing and extra/different methods for prompts 3 and 4 regarding the initial prompt results

| EID | Prompt 3 Missing | Prompt 3 Extra or different | Prompt 4 Missing | Prompt 4 Extra or different |
|---|---|---|---|---|
| 2-s2.0-84918834255 | X | Artifical Neural Network | X | X |
| 2-s2.0-84902106481 | X | X | X | X |
| 2-s2.0-84921642433 | X | X | X | X |
| 2-s2.0-84948392681 | X | Segmentation, Filtering, Localization, Learning, Non-Learning, ABCD Characteristics | X | X |
| 2-s2.0-84944318438 | X | Image processing | X | Image processing |
| 2-s2.0-84983425726 | X | CUDA programming framework, C++ programming language | X | X |
| 2-s2.0-84961903068 | X | Preprocessing, Segmentation, Feature Extraction, Classification | X | X |
| 2-s2.0-84988890309 | Sparse Coding | Pre-processing, Segmentation, Classification, SIFT, Hue, Opponent Angle Histograms, RGB Intensities, Dictionary Learning | X | X |
| 2-s2.0-84962792167 | X | Supervised learning | Supervised classification | X |
| 2-s2.0-84994285804 | X | Iterative Dilation Method, Illumination Compensation | X | X |
| 2-s2.0-84963804529 | LIPU | Machine Learning, Ordinal Classification | X | X |
| 2-s2.0-84976426664 | X | Predictive Modelling | X | X |
| 2-s2.0-85018542177 | X | Feature Extraction | Shape Characterization, Color and Texture Features | Feature Extraction |
| 2-s2.0-85019985302 | X | X | X | X |
| 2-s2.0-85039989899 | X | Image Processing, Computer Vision | X | X |
| 2-s2.0-85039912655 | X | Image Processing Software, Learning Program | X | X |
| 2-s2.0-85042474953 | X | X | X | X |
| 2-s2.0-85050502789 | X | Feature selection, Sequential backward selection, Image processing, Feature Extraction | X | Image Processing, Feature Extraction, Segmentation, Sequential Backward Selection, Feature Selection |

**Table 9** (continued)

| EID | Prompt 3 | | Prompt 4 | |
|---|---|---|---|---|
| | Missing | Extra or different | Missing | Extra or different |
| 2-s2.0-85042181700 | Support Vector Machine, Random Forest, Neural Network, Fast Discriminative MixedMembership-Based Naive Bayesian Classifiers | X | X | X |
| 2-s2.0-85074209854 | X | Dull-Razor software, Image Processing, Automatic Segmentation, Basic Statistical Method | X | X |
| 2-s2.0-85048760650 | X | Decision Tree | X | X |
| 2-s2.0-85060645376 | X | Statistical estimation | X | X |
| 2-s2.0-85059759729 | X | Image Processing | X | image processing techniques |
| 2-s2.0-85073726145 | Deep Learning | Natural Language Processing, Logics-based Systems | X | Deduction, Induction, Abduction, Radiomics, Natural Language Processing, Logics Based Systems |
| 2-s2.0-85081629940 | X | Cross-Validation, Image Processing, Artificial Intelligence | X | Artificial Intelligence |
| 2-s2.0-85091193990 | Artificical Intelligence | X | X | X |
| 2-s2.0-85099879408 | X | X | X | X |
| 2-s2.0-85086182046 | X | X | X | X |
| 2-s2.0-85082062276 | X | AI, Machine Learning | X | AI |
| 2-s2.0-85084500870 | X | Machine learning, Natural language processing, Computer vision | X | X |
| 2-s2.0-85093943094 | X | Receiver-Operating -Characteristic (ROC) curve | X | X |
| 2-s2.0-85123369429 | Image Segmentation | Path Direct Connection, Dropout Direct Connection, Conv Direct Connection, Constant Scale | Image Segmentation | Artifical Intelligence |
| 2-s2.0-85107175102 | X | Deep Learning | X | Deep Learning |
| 2-s2.0-85123014841 | X | CNN, Image Processing, Segmentation | X | X |
| 2-s2.0-85096782074 | X | tissue-type classification algorithm, nucleus detection | X | tissue-type classification algorithm, immunohistochemistry (IHC) |
| 2-s2.0-85108124165 | X | X | X | X |

**Table 9** (continued)

| EID | Prompt 3 | | Prompt 4 | |
|---|---|---|---|---|
| | Missing | Extra or different | Missing | Extra or different |
| 2-s2.0-85116348283 | X | X | X | X |
| 2-s2.0-85119477680 | X | X | X | X |
| 2-s2.0-85135422352 | X | X | X | X |
| 2-s2.0-85121330341 | X | Bidimensional Segmentation, Dimensionality Reduction | X | X |
| 2-s2.0-85125337919 | X | Deep Learning, Receiver Operating Characteristic (ROC), Fivefold Cross-Validation | X | Deep Learning, Receiver Operating Characteristic (ROC) |
| 2-s2.0-85124261031 | X | U-Net++ | U-Net | U-Net++ |
| 2-s2.0-85135382796 | X | Computed Tomography (CT), Dice similarity coefficient (DSC), Hausdorff distance transform (DT) | X | X |
| 2-s2.0-85131873486 | X | Frame-based contrastenhanced T1-weighted MR images, synthetic CT (sCT) | X | X |
| 2-s2.0-85124549532 | X | X | X | X |
| 2-s2.0-85134588716 | Deep Image-to-Image Network (DI2IN) | Dice Similarity Coefficient (DSC), Hausdorff Distance (HD95) | X | X |
| 2-s2.0-85144032371 | X | X | X | X |
| 2-s2.0-85146428707 | X | X | X | X |
| 2-s2.0-85146500215 | X | Patternnet, MATLAB | X | X |
| 2-s2.0-85148402129 | Machine Learning Algorithms | X | X | X |
| 2-s2.0-85159074452 | X | X | X | X |
| 2-s2.0-85131537962 | Artificial Intelligence | Transfer Learning, Image Data, Handcrafted Lesion Features, Patient-Centric Metadata, Multi-Input Single-Output (MISO) Model, Evaluation Metrics | Artificial Intelligence, Deep Learning | Transfer Learning |
| 2-s2.0-85152350497 | X | X | X | X |
| 2-s2.0-85147848611 | X | X | X | X |
| 2-s2.0-85148874931 | Artificial Intelligence | Image processing techniques, Wavelet transform | X | X |

**Table 10** Missing and extra/different methods for prompts 5 and 6 regarding the initial prompt results

| EID | Prompt 5 | | Prompt 6 | |
| --- | --- | --- | --- | --- |
| | Missing | Extra or different | Missing | Extra or different |
| 2-s2.0-84918834255 | X | X | X | X |
| 2-s2.0-84902106481 | X | X | X | X |
| 2-s2.0-84921642433 | X | X | X | X |
| 2-s2.0-84948392681 | X | Segmentation, Filtering, Localization, Learning, Non-Learning | X | X |
| 2-s2.0-84944318438 | X | Image processing | X | X |
| 2-s2.0-84983425726 | X | CUDA programming framework, C++ programming language | X | X |
| 2-s2.0-84961903068 | X | Preprocessing, Segmentation, Feature Extraction, Classification | X | X |
| 2-s2.0-84988890309 | Sparse Coding | Pre-processing, Segmentation, Classification, SIFT, Hue, Opponent Angle Histograms, RGB Intensities, Dictionary Learning | X | SIFT, Hue, Opponent Angle Histograms, RGB Intensities |
| 2-s2.0-84962792167 | X | X | Supervised classification, Multi-Layer Feed-forward Neural Network, Genetically Optimized Fuzzy C-means clustering | X |
| 2-s2.0-84994285804 | X | Iterative Dilation, Illumination Compensation, Feature Vector | X | Iterative Dilation Method |
| 2-s2.0-84963804529 | X | Machine Learning | X | X |
| 2-s2.0-84976426664 | X | Dice Similarity Coefficient (DSC) | X | Predictive Modelling |
| 2-s2.0-85018542177 | X | Feature Extraction | X | X |
| 2-s2.0-85019985302 | X | X | X | X |
| 2-s2.0-85039989899 | X | Image Processing, Computer Vision | X | Image Processing |
| 2-s2.0-85039912655 | X | Image Processing Software, Learning Program | X | Image Processing Software, Learning Program |
| 2-s2.0-85042474953 | X | X | X | X |
| 2-s2.0-85050502789 | X | Segmentation, Sequential backward selection, Image processing, Feature Extraction | X | Image Processing |

Table 10 (continued)

| EID | Prompt 5 | | Prompt 6 | |
|---|---|---|---|---|
| | Missing | Extra or different | Missing | Extra or different |
| 2-s2.0-85042181700 | X | X | Support Vector Machine, Random Forest, Neural Network, Fast Discriminative MixedMembership–Based Naive Bayesian Classifiers | X |
| 2-s2.0-85074209854 | X | Dull-Razor software, Image Processing, Automatic Segmentation, Basic Statistical Method | X | X |
| 2-s2.0-85048760650 | X | Decision Tree | X | Decision Tree |
| 2-s2.0-85060645376 | X | Statistical estimation | X | X |
| 2-s2.0-85059759729 | X | Image Processing Techniques | X | Image Processing |
| 2-s2.0-85073726145 | X | Natural Language Processing, Logics-based Systems | X | Deduction, Induction, Abduction, Radiomics, Natural Language Processing, Logics Based Systems |
| 2-s2.0-85081629940 | X | Cross-Validation, Image Processing, Artificial Intelligence | X | X |
| 2-s2.0-85091193990 | X | Image Processing | X | Deep Neural Network |
| 2-s2.0-85099879408 | X | X | X | X |
| 2-s2.0-85086182046 | X | X | X | X |
| 2-s2.0-85082062276 | X | X | X | X |
| 2-s2.0-85084500870 | X | Convolutional Neural Networks (CNNs), Generative Adversarial Networks (GANs), Recurrent Neural Networks (RNNs) | X | X |
| 2-s2.0-85093943094 | X | Receiver-Operating -Characteristic (ROC) curve | X | X |
| 2-s2.0-85123369429 | Image Segmentation | Path Direct Connection, Dropout Direct Connection, Conv Direct Connection, Constant Scale | Image Segmentation | Path Direct Connection, Dropout Direct Connection, Conv Direct Connection, Constant Scale |
| 2-s2.0-85107175102 | X | Deep Learning | X | X |
| 2-s2.0-85123014841 | X | X | X | X |

Table 10 (continued)

| EID | Prompt 5 | | Prompt 6 | |
|---|---|---|---|---|
| | Missing | Extra or different | Missing | Extra or different |
| 2-s2.0-85096782074 | X | tissue-type classification algorithm, nucleus detection, Computational pathology | X | tissue-type classification algorithm, nucleus detection, immunohistochemistry (IHC) |
| 2-s2.0-85108124165 | X | X | X | X |
| 2-s2.0-85116348283 | X | Dice similarity coefficient (DSC), Hausdorff distance (HD) | X | X |
| 2-s2.0-85119477680 | X | X | X | X |
| 2-s2.0-85135422352 | X | Lymph Node Recognition Algorithm | X | X |
| 2-s2.0-85121330341 | X | Bidimensional Segmentation, Radiomic Features, Dimensionality Reduction, Class Balancing, 10-Fold Cross-Validation, McNemar's Test | X | X |
| 2-s2.0-85125337919 | X | Receiver Operating Characteristic (ROC), Area Under the Receiver Operating Characteristic (AUROC) | X | Deep Learning, Receiver Operating Characteristic (ROC), Artificial Intelligence (AI) |
| 2-s2.0-85124261031 | X | U-Net + + | X | U-Net + + |
| 2-s2.0-85135382796 | X | Dice similarity coefficient (DSC), 95% Hausdorff distance transform (DT) | X | auto-contours, manual contours, Dice similarity coefficient, Hausdorff distance transform |
| 2-s2.0-85131873486 | X | Frame-based contrastenhanced T1-weighted MR images, synthetic CT (sCT), Convolution with rCT (Conv-rCT) plan, Convolution with sCT (Conv-sCT) plan | X | X |
| 2-s2.0-85124549532 | X | X | X | X |
| 2-s2.0-85134588716 | X | Dice Similarity Coefficient (DSC), Hausdorff Distance (HD95) | X | X |
| 2-s2.0-85144032371 | X | X | X | X |
| 2-s2.0-85146428707 | X | X | X | X |
| 2-s2.0-85146500215 | X | Patternnet, MATLAB | X | Patternnet |
| 2-s2.0-85148402129 | Machine Learning Algorithms | Keras library | Machine Learning Algorithms | Artificial Intelligence |
| 2-s2.0-85159074452 | X | X | X | X |

Table 10 (continued)

| EID | Prompt 5 | | Prompt 6 | |
|---|---|---|---|---|
| | Missing | Extra or different | Missing | Extra or different |
| 2-s2.0-85131537962 | Artificial Intelligence, Deep Learning | Transfer Learning Image Data, Handcrafted Lesion Features, Patient-Centric Metadata, Multi-Input Single-Output (MISO) Model, Weighing Models | Artificial Intelligence, Deep Learning, EfficientNets, Artificial Neural Network, Majority Soft Voting | X |
| 2-s2.0-85152350497 | X | X | X | X |
| 2-s2.0-85147848611 | X | Improved Ant Colony Optimizer (LACOR) | X | X |
| 2-s2.0-85148874931 | X | Image Processing, Wavelet Transform | X | Image Processing Techniques, Wavelet Transform |

**Table 11** Classified methods

| Method | 2014 | | | 2015 | | | 2016 | | |
|---|---|---|---|---|---|---|---|---|---|
| | Papers | Rel Sum | Pop Sum | Papers | Rel Sum | Pop Sum | Papers | Rel Sum | Pop Sum |
| Class 1 | 2 | 1 | 3.3 | 2 | 1 | 1.4444 | 2 | 0 | 7.125 |
| Class 2 | 0 | 0 | 0 | 0 | 0 | 0 | 0 | 0 | 0 |
| Class 3 | 0 | 0 | 0 | 0 | 0 | 0 | 2 | 1 | 6.25 |
| Class 4 | 1 | 1 | 2.7 | 0 | 0 | 0 | 1 | 0 | 1.75 |
| SVM | 0 | 0 | 0 | 1 | 0 | 1.1111 | 1 | 0 | 1.75 |
| K-means | 0 | 0 | 0 | 1 | 0 | 0.6667 | 0 | 0 | 0 |
| KNN | 1 | 0 | 0.8 | 0 | 0 | 0 | 0 | 0 | 0 |
| Logistic regression | 0 | 0 | 0 | 0 | 0 | 0 | 2 | 0 | 12 |
| GLCM | 0 | 0 | 0 | 0 | 0 | 0 | 0 | 0 | 0 |
| Method | 2017 | | | 2018 | | | 2019 | | |
| | Papers | Rel Sum | Pop Sum | Papers | Rel Sum | Pop Sum | Papers | Rel Sum | Pop Sum |
| Class 1 | 4 | 2 | 7.8572 | 1 | 0 | 4 | 1 | 0 | 0.2 |
| Class 2 | 0 | 0 | 0 | 0 | 0 | 0 | 2 | 0 | 0.4 |
| Class 3 | 0 | 0 | 0 | 1 | 0 | 4 | 0 | 0 | 0 |
| Class 4 | 0 | 0 | 0 | 0 | 0 | 0 | 0 | 0 | 0 |
| SVM | 0 | 0 | 0 | 1 | 0 | 4 | 1 | 0 | 1.2 |
| K-means | 0 | 0 | 0 | 0 | 0 | 0 | 0 | 0 | 0 |
| KNN | 0 | 0 | 0 | 1 | 1 | 2 | 0 | 0 | 0 |
| Logistic regression | 0 | 0 | 0 | 0 | 0 | 0 | 0 | 0 | 0 |
| GLCM | 0 | 0 | 0 | 1 | 1 | 0 | 0 | 0 | 0 |
| Method | 2020 | | | 2021 | | | 2022 | | |
| | Papers | Rel Sum | Pop Sum | Papers | Rel Sum | Pop Sum | Papers | Rel Sum | Pop Sum |
| Class 1 | 3 | 3 | 7.75 | 1 | 0 | 4.6667 | 1 | 0 | 1.5 |
| Class 2 | 6 | 4 | 39 | 9 | 2 | 23.0001 | 8 | 1 | 9 |
| Class 3 | 0 | 0 | 0 | 0 | 0 | 0 | 1 | 0 | 8.5 |
| Class 4 | 0 | 0 | 0 | 0 | 0 | 0 | 0 | 0 | 0 |
| SVM | 1 | 1 | 0 | 1 | 1 | 0.3333 | 0 | 0 | 0 |
| K-means | 0 | 0 | 0 | 0 | 0 | 0 | 0 | 0 | 0 |
| KNN | 0 | 0 | 0 | 1 | 1 | 0.3333 | 0 | 0 | 0 |
| Logistic regression | 0 | 0 | 0 | 0 | 0 | 0 | 0 | 0 | 0 |
| GLCM | 0 | 0 | 0 | 1 | 0 | 4.6667 | 0 | 0 | 0 |
| Method | 2023 | | | All Time | | | | | |
| | Papers | Rel Sum | Pop Sum | Papers | Rel Sum | Pop Sum | | | |
| Class 1 | 2 | 0 | 2 | 19 | 7 | 39.8433 | | | |
| Class 2 | 13 | 5 | 4 | 38 | 12 | 75.4001 | | | |
| Class 3 | 0 | 0 | 0 | 4 | 1 | 18.75 | | | |
| Class 4 | 1 | 1 | 1 | 3 | 2 | 5.45 | | | |
| SVM | 0 | 0 | 0 | 6 | 2 | 8.3944 | | | |
| K-means | 1 | 0 | 0 | 2 | 0 | 0.6667 | | | |
| KNN | 0 | 0 | 0 | 3 | 2 | 3.1333 | | | |
| Logistic regression | 0 | 0 | 0 | 2 | 0 | 12 | | | |
| GLCM | 0 | 0 | 0 | 2 | 1 | 4.6667 | | | |

**Class 2 (Deep learning methods)**: Deep learning (×15; 2019, 0, 0.2; 2020×3, 1, 27.75; 2021×3, 1, 4.3334; 2022×3, 1, 3.5; 2023×5, 1, 2), Generative Adversarial Network (GAN) (×2; 2019, 0, 0.2; 2020, 1, 3.75), ResNet (×1; 2020, 1, 3.75), ResNet50 (×1; 2021, 0, 4.6667), AlexNet (×2; 2020, 1, 3.75; 2021, 0, 4.6667), U-Net (×2; 2021, 1, 0; 2022, 0, 1.5), Convolutional Neural Network (CNN) (×4; 2021, 0, 4.6667; 2022, 0, 2; 2023×2, 1, 0), 2D U-Net (×1; 2021, 0, 2.3333), 3D U-Net (×1; 2021, 0, 2.3333), Deep Reinforcement Learning (DRL) (×1; 2022, 0, 1), Convolutional Encoder-Decoder Architecture (×1; 2022, 0, 1), Convolution algorithm (×1; 2022, 0, 0), Deep Convolutional Neural Network (DCNN) (×1; 2023, 0, 0), NASNetLarge (×1; 2023, 1, 0), DenseNet169 (×1; 2023, 1, 0), InceptionResNetV2 (×1; 2023, 1, 0), EfficientNets (×1; 2023, 0, 2), Conditional Generative Adversarial Network (cGAN) (×1; 2023, 0, 0).

**Class 3 (Tree-based methods)**: Random Forest (×2; 2016, 1, 2.125; 2018, 0, 4), Decision Trees (×1; 2016, 0, 4.125), Extra Trees Classifier (×1; 2022, 0, 8.5).

**Class 4 (Optimization methods)**: Genetic Algorithm (×1; 2014, 1, 2.7), Sequential Minimal Optimization (SMO) (×1; 2016, 0, 1.75), Ant Colony Optimization (ACO) (×1; 2023, 1, 1).

The cases are counted where the same method is used between 2014–2023, and all time. Relevancy and popularity sums are calculated for a specific method regarding the related articles. In other words, the first column ("Papers") states how many articles use the method in total. The second and third columns show the sum of relevancy and popularity values for these articles, respectively.

If all the time is considered, class 1, class 2, class 3, class 4, "K-nearest neighbors (KNN)", "support vector machine (SVM)", "K-means", "grey level co-occurrence matrix (GLCM)" and "logistic regression" are the ones that are mentioned in at least 2 articles. Sorting the total number of papers using these methods from largest to smallest is as follows:

**Papers**: class 2 > class 1 > "SVM" > class 3 > class 4 = "KNN" > "K-means" = "logistic regression" = "GLCM"
The relevancy values for all times are sorted as
**Relevancy**: class 2 > class 1 > class 4 = "SVM" = "KNN" > class 3 = "GLCM" > "K-means" = "logistic regression".
On the other hand, the sorting of popularity values for all time is given below and it indicates the highest value belongs to class 2.
**Popularity**: class 2 > class 1 > class 3 > "logistic regression" > "SVM" > class 4 > "GLCM" > "KNN" > "K-means".

From the above methods, it is seen that the number of implementing, and popularity trends of class 1 and class 2 have been increasing over the years. For this reason, tests can be started with AI methods in these classes in a similar problem domain.

**Abbreviations** **ABC**: Artificial bee colony; **ACO**: Ant colony optimization; **AEC**: Architecture, engineering and construction; **AI**: Artificial intelligence; **ANN**: Artificial neural network; **API**: Application programming interface; **BERT**: Bidirectional encoder representation from transformers; **BiLSTM**: Bidirectional long short-term memory; **BoW**: Bag of words; **BPNN**: Back propagation neural network; **BTM**: Bagging tree model; **CAI-Net**: Cloud attention intelligent network; **cGAN**: Conditional generative adversarial network; **CLAHE**: Contrast limited adaptive histogram equalization; **CNN**: Convolutional neural network; **DBSCAN**: Density-based spatial clustering of applications with noise; **DBN**: Deep belief network; **DCNN**: Deep convolutional neural network; **DeepSORT**: Deep simple online real-time tracking; **DNN**: Deep neural network; **DOI**: Digital object identifier; **DRL**: Deep reinforcement learning; **DSN**: Deeply supervised nets; **EID**: Electronic identifier; **ELM**: Extreme machine learning; **FCN**: Fully convolution networks; **FN**: False negative; **FP**: False positive; **GAN**: Generative adversarial network; **GCN**: Graph convolutional network; **GLCM**: Gray level co-occurrence matrix; **GNN**: Graph neural network; **GPT**: Generative pre-trained transformers; **GRU**: Gated recurrent unit; **GUI**: Graphical user interface; **IoT**: Internet of things; **KCNet**: Kernel-based canonicalization network; **KNN**: K-nearest neighbors; **LDA**: Latent Dirichlet allocation; **LSTM**: Long short-term memory; **MFFNN**: Multi-layer feed-forward neural network; **ML**: Machine learning; **MLP**: Multiple-layer perception; **NLP**: Natural language processing; **ORB**: Oriented fast and rotated brief; **PANN**: Paraconsistent artificial neural network; **PCA**: Principal component analysis; **PNN**: Probabilistic neural network; **PSO**: Particle swarm optimization; **RBM**: Restricted Boltzmann machine; **R-CNN**: Region-based convolutional neural network; **RNN**: Recurrent neural network; **SARBOLD-LLM**: Solution approach recommender based on literature database-large language model; **SIFT**: Scale-invariant feature transform; **SMO**: Sequential minimal optimization; **SSD**: Single shot detector; **SURF**: Speeded up robust features; **SVC**: Support vector classifier; **SVM**: Support vector machine; **SVR**: Support vector regression; **TF-IDF**: Term frequency-inverse document frequency; **TP**: True positive; **UAV**: Unmanned aerial vehicle; **XAI**: Explainable artificial intelligence; **XGBoost**: EXtreme Gradient Boosting; **YOLO**: You only look once

**Acknowledgements** All authors read and agreed to the published version of the manuscript.

**Author's Contribution** Deniz Kenan Kılıç: Conceptualization, Methodology, Software, Validation, Formal analysis, Investigation, Data curation, Writing – original draft, Writing – review & editing, Visualization. Alex Elkjær Vasegaard: Conceptualization, Methodology, Software, Validation, Formal analysis, Investigation, Data curation, Writing – original draft, Writing – review & editing, Visualization. Aurélien Desoeuvres: Conceptualization, Methodology, Software, Validation, Formal analysis, Investigation, Data curation, Writing – original draft, Writing – review & editing, Visualization. Peter Nielsen: Conceptualization, Methodology, Validation, Investigation, Data curation, Writing – review & editing, Supervision.

**Funding** No funding was received for this work.

**Data Availability** The data presented in this study are available upon request from the corresponding author.

## Declarations

**Ethics Approval and Consent to Participate**  Not applicable.

**Consent for Publication**  Not applicable.

**Competing Interest**  The authors declare that they have no known competing financial interests or personal relationships that could have appeared to influence the work reported in this article.

**Open Access**  This article is licensed under a Creative Commons Attribution 4.0 International License, which permits use, sharing, adaptation, distribution and reproduction in any medium or format, as long as you give appropriate credit to the original author(s) and the source, provide a link to the Creative Commons licence, and indicate if changes were made. The images or other third party material in this article are included in the article's Creative Commons licence, unless indicated otherwise in a credit line to the material. If material is not included in the article's Creative Commons licence and your intended use is not permitted by statutory regulation or exceeds the permitted use, you will need to obtain permission directly from the copyright holder. To view a copy of this licence, visit http://creativecommons.org/licenses/by/4.0/.

**Publisher's Note** Springer Nature remains neutral with regard to jurisdictional claims in published maps and institutional affiliations.